\documentclass[11pt]{article}

\usepackage[T1]{fontenc}
\usepackage[utf8]{inputenc}
\usepackage{times}
\usepackage{amsmath, amsfonts}
\usepackage{graphicx}
\usepackage{float}
\usepackage{array}
\usepackage{tabularx}
\usepackage{rotating}
\usepackage{fancyhdr}
\usepackage{enumitem}
\usepackage{hyperref}
\usepackage{url}
\usepackage[numbers]{natbib}

\usepackage{tikz}
\usetikzlibrary{arrows.meta, positioning, shapes.geometric, arrows, shapes}
\usepackage{ragged2e}
\usepackage{framed}
\usepackage{comment}
\usepackage{footnote}
\usepackage{ulem}
\usepackage{svg}
\usepackage{todonotes}
\usepackage{xurl} 
\usepackage[title,titletoc]{appendix}

\usepackage[margin=1in]{geometry}

\title{
    \centering
    {\Huge \bfseries AI-Driven Control \\[0.5em]
    of Bioelectric signalling \\[0.5em]
    for Real-Time Topological Reorganization of Cells}
}

\author{
    \textbf{Gonçalo Carvalho} \\
    \textit{goncalo.horacarvalho@gmail.com}
}

\date{\today}

\setlist{nolistsep}

\begin{document}

\begin{titlepage}
    \centering
    {\scshape \Large \hspace{1mm} \\ \small October~~~||~~~2024 \vspace{1cm} \\ \Huge AI-Driven Control \\ of Bioelectric signalling \\ for Real-Time \\ Topological Reorganization \\ of Cells}
    \vspace{1.7cm} \\
    {\scshape Zurich, Switzerland \\ }
    \vspace{1.3cm}
    
    {\protect\NoHyper
    Gonçalo Carvalho,\footnote{goncalo.horacarvalho@gmail.com}
    \protect\endNoHyper
    }
    
    \vspace{2.5cm}
\end{titlepage}

\newpage
\tableofcontents
\newpage



\begin{abstract}
Understanding and manipulating bioelectric signaling could present a new wave of progress in developmental biology, regenerative medicine, and synthetic biology. Bioelectric signals, defined as voltage gradients across cell membranes caused by ionic movements, play a role in regulating crucial processes including cellular differentiation, proliferation, apoptosis, and tissue morphogenesis. Recent studies demonstrate the ability to modulate these signals to achieve controlled tissue regeneration and morphological outcomes in organisms such as planaria and frogs. However, significant knowledge gaps remain, particularly in predicting and controlling the spatial and temporal dynamics of membrane potentials (V\_mem), understanding their regulatory roles in tissue and organ development, and exploring their therapeutic potential in diseases.

In this work we propose an experiment using Deep Reinforcement Learning (DRL) framework together with lab automation techniques for real-time manipulation of bioelectric signals to guide tissue regeneration and morphogenesis. The proposed framework should interact continuously with biological systems, adapting strategies based on direct biological feedback.

Combining DRL with real-time measurement techniques - such as optogenetics, voltage-sensitive dyes, fluorescent reporters, and advanced microscopy - could provide a comprehensive platform for precise bioelectric control, leading to improved understanding of bioelectric mechanisms in morphogenesis, quantitative bioelectric models, identification of minimal experimental setups, and advancements in bioelectric modulation techniques relevant to regenerative medicine and cancer therapy. Ultimately, this research aims to utilize bioelectric signaling to develop new biomedical and bioengineering applications.
\end{abstract}

\section{Introduction}

\subsection{Bioelectric Signaling in Morphogenesis and Regeneration}
The intricate processes of cellular behavior, tissue formation, and morphogenesis, if uncovered, would fundamentally change developmental biology, regenerative medicine, and synthetic biology. Recent research has increasingly highlighted the pivotal role of bioelectric signaling, gene expression dynamics, and computational modeling in understanding and manipulating these processes.

McLaughlin and Levin \cite{McLaughlin2018} explored bioelectric signaling in regeneration, demonstrating that specific voltage modulations can direct tissue structure outcomes. Their experiments with \textit{Xenopus} tadpoles involved manipulating transmembrane potentials (Vmem) using ion channel overexpression and optogenetic tools to induce or inhibit limb regeneration. Pharmacological agents such as ouabain were used to influence ion flux, while optogenetic tools allowed precise, spatially resolved control over bioelectric states through light-induced activation of specific ion channels. Live imaging tracked morphological changes, and genetic markers identified differentiation in response to bioelectric alterations. Results showed that targeted Vmem modulations led to significant changes in regeneration patterns, promoting or inhibiting cell proliferation and differentiation as needed. For example, hyperpolarising cells promoted directed differentiation, while depolarizing states increased proliferation. These findings support the concept of a bioelectric code as an instructive signal for tissue organization and regenerative processes, further validating the therapeutic potential of bioelectric modulation in regenerative medicine by providing positional information crucial for proper morphogenesis.

Practical applications of bioelectric principles are further exemplified by Houpu Li et al. \cite{Li2023}, who design, fabricate, and test a wearable bioelectronic bandage capable of delivering fluoxetine to wounds to enhance healing. The methodology involved designing a bioelectronic bandage comprising an ion pump delivery module and a controller module. The ion pump, fabricated using polydimethylsiloxane (PDMS), utilized hydrogel-filled capillaries as ion-selective membranes to allow positively charged fluoxetine ions to pass while blocking other ions. The controller module included a programmable printed circuit board (PCB) managing precise fluoxetine dosage through electrophoretic force. In vivo testing on C57B6 male mice demonstrated that fluoxetine delivery from the bioelectronic bandage significantly increased re-epithelialization by 39.9\% and reduced the M1/M2 macrophage ratio by 27.2\%, indicating enhanced wound progression and reduced inflammation. These results illustrate the therapeutic potential of programmable bioelectric devices in medical applications, showcasing the integration of bioelectric principles with wearable technology for targeted drug delivery and improved healing outcomes.

Further supporting the role of bioelectric signaling in regeneration, Beane et al. \cite{Beane2013} investigate how H,K-ATPase ion transporters influence tissue remodeling and organ scaling in planarian regeneration. Their experiments with \textit{Schmidtea mediterranea} planarians involved RNA interference (RNAi) to inhibit H,K-ATPase function, resulting in regenerates with disproportionately large pharynges and shrunken heads. Apoptosis assays revealed that H,K-ATPase inhibition led to reduced apoptosis during the later stages of regeneration, disrupting normal tissue remodeling and causing pre-existing tissues to retain their original size and shape instead of adjusting proportionally to the regenerating body. Membrane potential measurements and intracellular pH recordings confirmed changes in cellular ion states due to H,K-ATPase RNAi. Immunofluorescence and in situ hybridization techniques further demonstrated disruptions in tissue remodeling markers, underscoring the essential role of bioelectric regulation in maintaining tissue homeostasis and proper organ scaling during regenerative processes.

In the realm of collective behavior and stress management, Shreesha and Levin \cite{SHREESHA2024150396} present a computational model where stress sharing among cells facilitates coordinated morphogenesis. The model simulates a virtual embryo on a two-dimensional grid, where each cell can exhibit one of two states representing different cell types. Cells are randomly arranged initially, forming a scrambled configuration that deviates from the target pattern. Each cell experiences stress quantified as a binary signal indicating whether it is in its correct target position. In stress-sharing configurations, stressed cells can diffuse their stress to neighbors, creating a shared stress environment that encourages collective movement toward target positions. Genetic algorithms evolved these virtual embryos over generations, assessing fitness based on the L2 distance from the target pattern. Results showed that stress-sharing embryos achieved higher fitness faster than those without stress-sharing or with hardwired configurations. Stress-sharing increased the movement range for individual cells, allowing them to resolve their positions more effectively and achieve coherent tissue organization. 

Hansali et al. \cite{Hansali2024} examine the influence of bioelectrical patterns on morphogenesis through evolutionary simulations and empirical validation in planarian regeneration. Utilizing a neural cellular automata (NCA) model, the study investigates how different bioelectric encodings - direct, indirect, and binary triggers - contribute to developmental outcomes. The methodology involved simulating 9x9 cell grids representing artificial organisms, each controlled by a neural network optimized via a genetic algorithm over 250 generations. Target morphologies, such as the "Tadpole" and the "French Flag," inspired by actual biological morphogenetic processes, were used to assess the effectiveness of different encoding strategies. Results demonstrated that direct pattern organisms reliably achieved target morphologies even after bioelectric resetting due to their linear encoding. Indirect pattern organisms exhibited emergent robustness against partial bioelectric disruptions but struggled with complete resets, while binary trigger organisms were highly sensitive to precise timing of bioelectric cues, failing morphogenesis under altered conditions. Empirical validation through SSRI experiments in planarian regeneration confirmed that SSRIs disrupt bioelectric interpretation, resulting in variable morphologies and bistable anatomical states. This underscores the critical role of bioelectric signaling in developmental regulation and the robustness of indirect encoding strategies against environmental perturbations.

\subsection{Synthetic Biology and Multicellularity}
The challenges in synthetic multicellularity are comprehensively addressed by \cite{Sole2024}, who categorize synthetic multicellular systems into circuits, programmable assemblies, and synthetic morphologies. The methodology spans genetic modification to create complex cellular responses, such as Boolean logic gates and pattern formation circuits, as well as manipulating cell adhesion properties to influence spatial organization within cell collectives. Experiments in synthetic multicellular circuits involved designing multicellular consortia where cellular strains communicate through chemical signals in liquid media or agar plates, resulting in stable configurations or patterns under controlled conditions. Programmable synthetic assemblies utilized differential adhesion models to drive self-organized cellular aggregates, achieving predictable spatial arrangements based on specific initial conditions. Synthetic morphology experiments, including the development of Xenobots and Anthrobots, emphasized the emergence of functionalities like memory and homeostasis without genetic editing, showcasing the balance between engineered predictability and emergent biological behaviors. The results highlight both the successes and inherent challenges in scaling synthetic multicellular systems, pointing to the necessity for novel methodologies to manage complexity and unpredictability in larger-scale constructs.

Advancements in bioinformatics and network analysis have further elucidated the connections between bioelectric signaling and disease. Pio-Lopez and Levin \cite{PioLopez2024} introduce MultiXVERSE, a universal multilayer network embedding method applied to investigate the causal link between GABA neurotransmitters and cancer. By constructing a multiplex-heterogeneous network that combines drug, disease, and gene interactions, MultiXVERSE employs Random Walks with Restart (RWR) for node similarity calculations and link prediction via a Random Forest classifier. The study involved creating a multilayer network incorporating protein-protein interactions, drug interactions from sources like DrugBank, and disease networks representing symptom similarity and comorbidity relationships. Experimental validation using \textit{Xenopus laevis} embryos exposed to the GABA agonist muscimol demonstrated that fluoxetine delivery via a wearable bioelectronic bandage significantly enhanced wound healing, evidenced by a 39.9\% increase in re-epithelialization and a 27.2\% reduction in the M1/M2 macrophage ratio. These findings suggest that bioelectric states, modulated by neurotransmitters like GABA, play a significant role in tumorigenesis and cancer progression, offering potential avenues for therapeutic interventions through bioelectric modulation.

Exploring the evolutionary aspects of multicellular intelligence, Hartl, Risi, and Levin \cite{Hartl2024} investigate how different levels of cellular competency within a multi-scale competency architecture (MCA) affect the evolutionary process. Their in-silico model simulates morphogenesis using neural cellular automata (NCAs), where cellular agents evolve to form target patterns. The methodology involves setting up NCAs on an 8x8 grid to form predefined patterns, such as a Czech flag, through coordinated cell state updates. Each cell contains an artificial neural network (ANN) that processes information from neighboring cells and proposes state changes. Key competency parameters, including decision-making probability (PD) and redundancy in decision pathways (R), are adjusted to simulate various levels of cellular reliability and computational capacity. The evolutionary fitness of each NCA is measured based on the accuracy and stability of forming the target pattern, with cells experiencing noise to simulate environmental challenges. Results indicate that higher competency levels accelerate evolutionary success and enhance the robustness and adaptability of morphogenetic processes. Additionally, NCAs with multi-scale competency encoding demonstrated better generalization to new environmental conditions compared to direct encoding models, suggesting that cellular-level intelligence significantly contributes to the adaptability and resilience of developing structures.

The concept of memory as a dynamic and adaptive process is further explored by Levin \cite{Levin2024}, who posits that biological memories are not static repositories but are continuously reinterpreted to maintain functional relevance during significant physiological changes. Through conceptual analysis and comparative studies, Levin illustrates how memory engrams adapt to new biological contexts, such as metamorphosis in caterpillars and planarian regeneration. For instance, during metamorphosis, memories associated with a caterpillar’s sensory-motor systems are reformatted to remain functionally relevant for the butterfly's needs, despite drastic physical and behavioral transformations. Similarly, planaria can transfer memories through body regeneration, revealing that specific memory cues adapt across different biological states. This phenomenon, termed "mnemonic improvisation," is central to understanding intelligence, as memory supports flexible, real-time problem-solving and continuity of self. Levin’s perspective offers a framework for understanding intelligence as an emergent property of dynamic memory processes that evolve across different biological scales.

Investigating the emergence of higher-level agency, Tissot et al. \cite{Tissot2024} examine how phase synchronization among individual decision cycles in a decentralized system can lead to the emergence of higher-level agency. Their computational model simulates collective behaviors of individual units that engage in synchronized decision-making, achieving forms of collective action that surpass the capabilities of individual units acting alone. Each unit operates within a modular system, interacting more frequently with intra-module units than with inter-module units. The decision cycles are governed by oscillatory functions where phase adjustments occur based on the alignment of decisions within each module. Results demonstrate that synchronization leads to spontaneous collective shifts from local to global optima, overcoming individual-level stress barriers. After synchronization, modules coordinate responses across the entire system, achieving states with minimized collective stress. These findings suggest that synchronization mechanisms are fundamental to the emergence of collective intelligence in decentralized biological systems, enabling higher-level problem-solving competencies.

In cancer research, Kofman and Levin \cite{Kofman2024} perform a meta-analysis evaluating pharmacological agents targeting ion channels as potential cancer therapies. Their comprehensive literature search across databases like IUPHAR, DrugBank, and PubMed identified various ion channel-targeting drugs, categorizing them based on their cancer-promoting or inhibiting effects. The analysis revealed that numerous ion channel drugs exhibit significant cancer-modifying properties across multiple cancer types, including breast, liver, brain, and ovarian cancers. Voltage-gated sodium and potassium channels frequently appeared in both pro- and anti-cancer categories, suggesting that the bioelectric state of cells can significantly influence tumor progression and metastasis. Some drugs demonstrated dual effects, indicating that factors such as timing, dosage, or biological context might influence their impact on cancer phenotypes. These findings highlight the therapeutic potential of bioelectric modulation in cancer treatment strategies, offering new targets for intervention beyond traditional genetic approaches.

Zhang, Goldstein, and Levin \cite{zhang2023classicalsortingalgorithmsmodel} explore morphogenetic processes through the application of classical sorting algorithms. By implementing sorting algorithms such as Bubble Sort, Insertion Sort, and Selection Sort in a decentralized, cell-view model, they simulate self-sorting cellular arrays capable of handling error-prone conditions through emergent behaviors like clustering and delayed gratification. The methodology involved treating each element in an array as an autonomous agent capable of viewing its neighboring elements and deciding whether to swap based on local conditions. In addition to normal, fully-functional cells, the study introduced "frozen" cells that could either initiate no swaps or be immovable by others, simulating unreliability as seen in biological media. Results demonstrated that cell-view algorithms maintained sorting efficiency and robustness even in the presence of unreliable cells, showcasing emergent problem-solving behaviors in minimalistic systems. These findings indicate that algorithmic models can effectively mimic biological self-organization and problem-solving capabilities, offering insights into the decentralized control mechanisms inherent in biological morphogenesis.

Theoretical frameworks for self-organization during prenatal development are advanced by Ciaunica et al. \cite{Ciaunica2023}, who apply the concept of Markov blankets to model the maternal-fetal relationship during human prenatal development. Their study emphasizes the interplay between maternal and fetal immune systems, conceptualizing pregnancy as a dynamic process where two immune systems operate as self-organizing entities with distinct yet interconnected Markov blankets. The placenta serves as a boundary regulating nutrient and signal exchange, maintaining separate yet coordinated environments for both the mother and fetus. The methodology involved a detailed analysis of immune interactions, including the migration of immune cells like macrophages and natural killer cells through the placenta, contributing to the unique immune environment of pregnancy. The results highlight a theoretical framework where the Markov blanket formalism provides insights into the maternal-fetal relationship as an instance of nested selves, facilitating co-homeostasis and coordinated developmental processes. This perspective offers a novel understanding of biological self-organization, emphasizing the role of shared Markov blankets in maintaining the balance and stability of interconnected yet distinct biological systems during prenatal development.

Decentralized control mechanisms in microswimmers are investigated by Hartl, Levin, and Zöttl \cite{Hartl}, who demonstrate that artificial neural networks (ANNs) and genetic algorithms can optimize collective locomotion strategies in N-bead swimmers modeled after the Najafi-Golestanian (NG) microswimmer. The methodology involved constructing a virtual N-bead swimmer model where each bead acts as an independent agent capable of perceiving the states of its neighboring beads and applying forces accordingly. Using reinforcement learning techniques, the ANN parameters were evolved to maximize the swimmer’s speed, with the study testing swimmer sizes ranging from three to 100 beads. Results showed that the decentralized approach enabled robust and efficient swimming gaits, with locomotion strategies scaling effectively as the number of beads increased. Type B swimmers, in particular, displayed collective contractions reminiscent of caterpillar-like movement. Additionally, the evolved locomotion strategies demonstrated resilience to morphological changes and adaptability to different swimmer configurations without retraining, highlighting the potential of decentralized systems in achieving coordinated movement in autonomous microswimmers for biomedical applications.

Levin \cite{Michael2019} theorizes that bioelectric networks in multicellular systems establish cognitive boundaries that enable cells to function as unified cognitive entities. By integrating developmental biology, bioelectric signaling, and cognitive science, Levin posits that bioelectric gradients serve as computational mediums guiding multicellular coordination and pattern formation. The methodology involves a theoretical framework combining bioelectric signaling mechanisms, such as ion channels and gap junctions, with cognitive boundary concepts to model how cells collectively process information and regulate behavior. Examples from studies on \textit{Xenopus} embryos, planarians, and tadpole regeneration support the hypothesis that bioelectric states function as pre-patterns, directing cellular organization and overriding genetic instructions to achieve specific anatomical outcomes. The results indicate that bioelectric signaling underpins cognitive-like processes across biological scales, suggesting that the "self" emerges from bioelectric boundaries that define the spatio-temporal limits within which cells interact and maintain coherence as a unified system.

Lyon et al. \cite{Lyon2021} expand the understanding of cognitive processes in biology by redefining basal cognition to include sensory and information-processing mechanisms observed across diverse life forms. Their study examines cognitive behaviors in single-celled organisms, prokaryotes, and plants, focusing on capacities such as sensing, perception, memory, decision-making, and communication. By investigating mechanisms like quorum sensing in bacteria, cellular adhesion in amoebas, and bioelectrical signaling in plants, the authors argue that foundational cognitive capacities are evolutionarily conserved and precede the development of nervous systems. Experiments involved monitoring cellular responses to nutrient gradients and light stimuli, eliciting directional movement and adaptive behaviors without the presence of a nervous system. The results reveal that even simple organisms exhibit basic cognitive functions, suggesting a continuity of cognitive processes across biological complexity. This foundational understanding informs the study of more advanced cognitive systems and has potential applications in regenerative medicine and artificial intelligence, where principles of basal cognition can inspire novel approaches to problem-solving and adaptive behaviors.

In "Resting Potential, Oncogene-induced Tumorigenesis, and Metastasis: The Bioelectric Basis of Cancer in vivo," Lobikin et al. \cite{Lobikin2012} utilize \textit{Xenopus laevis} tadpoles to investigate how bioelectric signaling influences cancerous transformations. The methodology involved modulating transmembrane voltage potentials (Vmem) by depolarizing specific GlyCl-expressing instructor cells using pharmacological agents like ivermectin to open glycine-gated chloride channels. By manipulating extracellular chloride concentrations, controlled depolarization was induced, leading to melanocytes exhibiting metastatic-like transformations, including hyperproliferation, morphological changes, and invasive behavior. Additionally, exposure to the carcinogen 4-Nitroquinoline 1-oxide (4NQO) generated localized tumors, with bioelectric changes visualized using fluorescent voltage- and ion-reporter dyes. The study found that oncogene-induced tumors displayed elevated sodium levels, potentially useful for non-invasive diagnostics. Conversely, forced hyperpolarization of cells via expression of hyperpolarizing ion channels mitigated oncogene-induced tumorigenesis, demonstrating that bioelectric state modulation can suppress tumor formation. These findings highlight the active role of bioelectric signaling in cancer progression and metastasis, suggesting bioelectric modulation as a promising strategy for cancer therapeutics.

Pezzulo and Levin \cite{Pezzulo2016} advocate for top-down approaches in understanding and controlling complex biological systems, contrasting with traditional bottom-up, molecular-centric frameworks. Their study emphasizes the benefits of higher-level organizational strategies, akin to those in physics and engineering, for regulating complex processes such as pattern formation and regeneration. The methodology involves integrating concepts from information theory, control theory, and computational neuroscience to develop models that focus on system-wide goal states and dynamic regulation rather than precise molecular manipulation. Examples include the use of least-action principles in physics and control-theoretic models in neuroscience to achieve system stability and desired outcomes. In developmental biology, the authors propose that targeting large-scale anatomical goals can guide cellular behavior toward desired outcomes through bioelectric signaling and pattern memory. Studies on phenomena like planaria regeneration, where bioelectric interventions led to permanent anatomical changes, support the feasibility of top-down models. The results suggest that focusing on higher-level system organization can provide more effective control over complex biological processes, paving the way for innovative approaches in regenerative medicine and synthetic bioengineering by harnessing goal-directed, system-wide controls over molecular mechanisms.

\subsection{Questions}
Bioelectric signals in cells are, again, electrical potentials generated by the movement of ions (such as sodium, potassium, calcium, and chloride) across the cell membrane through ion channels and pumps. Bioelectric signals are represented by the voltage gradients across cell membranes ($V_{mem}$) and play an important role in developmental biology, governing processes such as tissue formation, wound healing, and organogenesis. Recent work by Michael Levin and others has demonstrated that it is possible to manipulate these bioelectric signals, showing the capacity to direct tissue regeneration in organisms like planaria and frogs \cite{Levin2014wh}. Yet, many questions remain unanswered, such as: \\

\small 
\begin{tabular}{| p{0.5cm} | p{13.5cm} |}
\hline
\textbf{No.} & \textbf{Question} \\
\hline
1 & How can we predict the spatial distribution of absolute $V_{mem}$ values (the specific numerical value of the membrane potential at a given location in millivolts) within a cell group, relative differences in $V_{mem}$ across cell borders, and time-dependent changes of $V_{mem}$ within cells? \\
\hline
2 & How do bioelectric signals regulate and control the anatomical growth and form of different tissues and organs? \\
\hline
3 & Can we identify and manipulate the prototypical morphological blueprint of complex tissues and organs without genetic intervention? \\
\hline
4 & What bioelectric mechanisms enable a collection of cells to function cohesively as an organ, maintaining homeostasis and structural stability? \\
\hline
5 & What are the decision-making capacities of cells, and what degrees of freedom do they possess? \\
\hline
6 & How do problematic cells, such as cancerous cells, override homeostatic mechanisms like cell-to-cell communication? \\
\hline
7 & Can artificially induced bioelectric signalling prevent cellular defection and enforce conformity within tissues? \\
\hline
8 & How many cells are required to effectively study these phenomena? \\
\hline
9 & How do cells compare and coordinate their bioelectric states across distances in a tissue, organ, or even the whole organism? \\
\hline
10 & How can we develop quantitative models of bioelectric circuits that reliably store stable patterning information during morphogenesis? \\
\hline
11 & What new synthetic biology tools will enable the top-down programming of bioelectric circuits for applications in regenerative medicine and cancer therapy? \\
\hline
12 & How can optogenetics be expanded to control stable $V_{mem}$ states in large, nonexcitable cell groups for better bioelectric manipulation? \\
\hline
13 & Can bioelectric circuits in nonneural cells store and process information in a manner similar to the way neural circuits store behavioural memory? \\
\hline
14 & How can new voltage reporters and techniques be developed for better in vivo modulation and observation of bioelectric states in real-time? \\
\hline
\end{tabular}

\newpage

As organismal complexity increases, predicting and optimizing bioelectric-driven morphogenesis 
becomes a significantly harder challenge. To help answer some of the questions above, this proposal seeks to address this challenge by developing a novel \textit{Deep Reinforcement Learning} (DRL) framework capable of interacting with real biological systems to directly control bioelectric signals in real-time. Unlike traditional computational models, this approach will allow for real-time adaptation by learning directly from biological feedback. In parallel, the data gathered from these experiments will be used to build computational models (i.e. digital twins) to better understand cellular organisation in more complex tissues, organs, and organisms.

\subsection{Research Objectives}
The specific objectives are:

\begin{itemize}
    \item \textbf{Objective 1}: Design and implement a DRL framework capable of learning optimal strategies for bioelectric signal manipulation to control cell proliferation, differentiation, migration, shape, apoptosis, and gene expression in order to guide overall tissue morphogenesis and regeneration. 
    \item \textbf{Objective 2}: Validate the effectiveness of the DRL framework through experiments with model organisms, starting with simple systems like yeast bacteria or planarians and scaling up to more complex organisms such as \textit{Drosophila melanogaster} (fruit flies). 
    \item \textbf{Objective 3}: Construct computational models (digital twins) of biological tissues based on experimental data, enhancing our understanding of cellular organization in complex tissues and organs. 
\end{itemize}

For Objective 2, for example, a study in replicability can be conducted following \cite{Pietak2016} where an open-source framework for simulating 2D computational multiphysics problems such as electrodiffusion, electro-osmosis, galvanotaxis, voltage-gated ion channels, gene regulatory networks, and biochemical reaction networks (e.g., metabolism) has been made available through \href{https://github.com/betsee/betse}{BETSE (BioElectric Tissue Simulation Engine) Github}. 

By achieving these objectives, this research seeks to advance our ability to control and direct morphogenetic processes, potentially leading to significant breakthroughs in regenerative medicine, cancer treatment, and synthetic biology. We are also not starting from scratch. 

\section{Background and Motivation}
\subsection{Current Understanding of Bioelectric Signals}

Bioelectric signals are fundamental to the regulation of cellular and developmental processes. They are generated by the movement of ions across cellular membranes, creating voltage gradients ($V_{\text{mem}}$) that influence cell behaviour. Recent research has begun to map the bioelectric patterns associated with various cellular states and developmental processes, revealing that these patterns play a crucial role in tissue formation, regeneration, and organogenesis \cite{Levin2009, Levin2014wh, adams2013endogenous}.

Just as the genome has been sequenced to understand genetic contributions to biology, efforts have been made to map bioelectric patterns in organisms. Techniques such as voltage-sensitive dyes and fluorescent reporters have allowed visualization of membrane potential distributions in living tissues. Studies have shown that specific bioelectric patterns correlate with particular developmental outcomes. For example, the formation of the anterior-posterior axis in \textit{Xenopus laevis} embryos is regulated by bioelectric cues \cite{adams2013endogenous, Pai2012}.

However, unlike the genome, the bioelectric "map" is dynamic and context-dependent. The spatial and temporal variability of bioelectric signals across different cell types and developmental stages makes comprehensive mapping challenging. Current maps are incomplete and often organism-specific, limiting the generalization of findings \cite{McCaig2005, Sundelacruz2009}.

Cells interpret bioelectric signals through voltage-gated ion channels, transporters, and voltage-sensitive signalling pathways. Changes in membrane potential can influence gene expression, cell proliferation, differentiation, migration, and apoptosis. For instance, depolarization of membrane potential has been shown to promote stem cell proliferation, while hyperpolarization can induce differentiation \cite{Levin2009, Levin2014wh, Blackiston2011}.

Despite these insights, the precise mechanisms by which cells transduce bioelectric signals into specific biochemical responses remain poorly understood. The complexity arises from the interplay between electrical, chemical, and mechanical signals within the cellular microenvironment. Additionally, the same bioelectric signal can elicit different responses depending on the cell type and its developmental context \cite{Levin2014wh}.

\subsection{Bioelectricity and its Role in Topological Control}

Bioelectricity, the result of ion flow across cellular membranes, acts as a fundamental signal that coordinates cell behaviour, gene expression, and tissue organization. These voltage gradients serve as a communication medium that enables cells to “know” their relative position within an organism, guiding processes like cell proliferation, differentiation, and morphogenesis \cite{Levin2014wh, Whited2019}. Levin's experiments with \textit{planaria} have demonstrated that by manipulating these bioelectric patterns, researchers can direct tissue outcomes  -  regenerating a planaria's head or a tail based solely on the bioelectric state \cite{Beane2011-yb}. These findings suggest that bioelectric signals function as a form of "code" that governs tissue identity, growth, and maintenance.

Additionally, bioelectric signals have been shown to play a role in broader topological manipulations. For instance, in \textit{Xenopus laevis} (the African clawed frog), bioelectric manipulations have been used to induce the formation of eyes in unconventional locations, illustrating the potential of bioelectric control to not only regenerate tissues but also alter and transform them in novel ways \cite{Levin2007-eb}. This points to the possibility that the bioelectric "fingerprint" behind an organ or tissue could be decoded, allowing for manipulation of $V_{mem}$ at a fundamental level, potentially enabling the full recovery of damaged organs, such as in cardiac arrest victims, complete overhaul of malfunctioning tissues like cancer, and perhaps even the transformation of tissues into entirely novel structures \cite{Davies2020}.

Based on the work of Souidi et al., we take the drosophila fly as our goal system for modelling heart disease and recovery induced by the proposed DRL system \cite{Souidi2021}.

\subsection{Biomedical and Engineering Implications}

The biomedical implications of cracking the bioelectric code are many. By understanding and controlling the high-level triggers of morphogenetic subroutines, it might seem like a giant leap in imagination to picture a moment in time when we'll be able to induce complex regenerative outcomes such as limb and organ regeneration, repair of birth defects, and even the prevention and reprogramming of diseases like cancer via bioelectric pathways. But these processes have been made possible to some extent already in vivo, both using human cells as well as in complex organisms, such as frogs \cite{Levin2014wh}.

Bioelectric manipulation has shown promise in modulating the tumor environment, potentially suppressing or redirecting cancerous cell behaviour \cite{Chernet2013}, while optogenetic control of voltage gradients has successfully triggered regeneration in non-regenerative species \cite{Whited2019}. Simulations, like those done with BETSE, allow us to predict and manipulate bioelectric patterns across tissues, aiding in the exploration and improvement over the control of cellular behaviour during morphogenesis \cite{Pietak2016}. Moreover, advances in synthetic biology have begun to explore self-assembling tissues with designed bioelectric circuits, paving the way for the engineering of new tissues and organs, as well as the development of novel therapeutic strategies for regenerative medicine, such as in victims of frostbite, and synthetic life \cite{Davies2020}.

Furthermore, the implications extend to the treatment of heart anomalies and the recovery from heart attacks. So by using bioelectric signalling to induce proper heart patterning, we can potentially correct structural defects and restore normal function in damaged tissues. For instance, by applying targeted bioelectric stimulation or using pharmacological agents to modulate ion channels in cardiac tissues, it may be possible to restore the electrical and structural integrity of the heart following an injury or anomaly. By manipulating bioelectric patterns, we can aim to guide stem cells or progenitor cells to differentiate into cardiomyocytes, facilitating the repair of damaged heart muscle and ultimately enhancing the heart's functional capacity.

\section{Theoretical Minimum to Model the Problem}
The objective is to model a complex biological system, from single cells to tissues and organs, involving various ion channels, pumps, and signalling pathways within each cell. The AI agent interacts with the system by adjusting the membrane potential ($V_{\text{mem}}$) through techniques such as voltage clamping or optogenetics. The ultimate goal is to control cellular behaviour by manipulating $V_{\text{mem}}$ to achieve desired biological outcomes.

\subsection{modelling of Cellular Components}

\begin{figure}[H]
    \centering
    \includegraphics[width=0.5\textwidth]{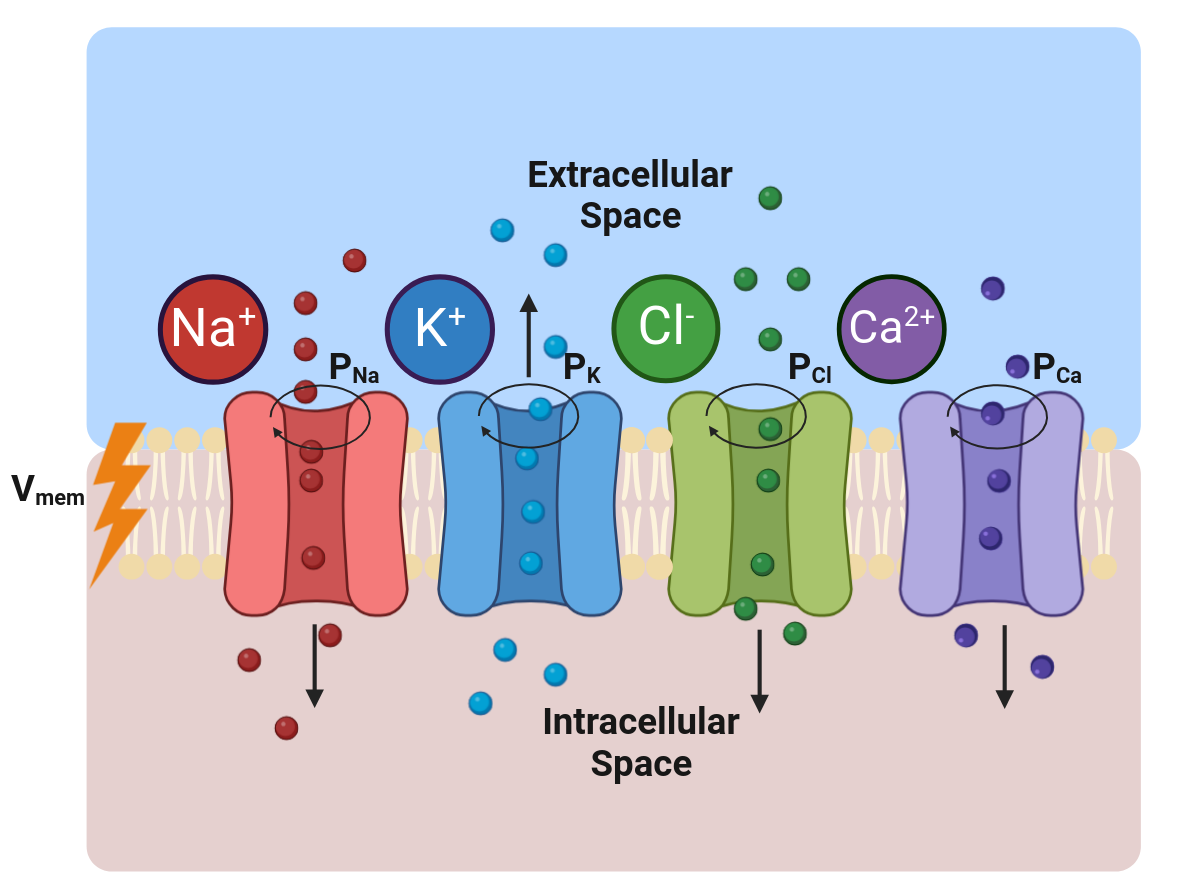} 
    \caption{Illustration of the cellular components used in the GHK equation.}
    \label{fig:GHK}
\end{figure}

The membrane potential dynamics are described by the Goldman-Hodgkin-Katz (GHK) equation when multiple ionic species contribute significantly to the membrane potential. The GHK voltage equation illustrated in Fig. \ref{fig:GHK} calculates the membrane potential $V$ based on the relative permeabilities and concentrations of the predominant ions:

\begin{equation} V = \frac{RT}{F} \ln\left( \frac{P_{\text{Na}} [\text{Na}^+]{_{out}} + P{\text{K}} [\text{K}^+]{_{out}} + P{\text{Cl}} [\text{Cl}^-]{_{in}}}{P{\text{Na}} [\text{Na}^+]{_{in}} + P{\text{K}} [\text{K}^+]{_{in}} + P{\text{Cl}} [\text{Cl}^-]{_{out}}} \right) \end{equation}

Here, $R$ is the universal gas constant, $T$ is the absolute temperature of the cell in Kelvin, $F$ is Faraday's constant corresponding to the amount of electricity that is carried by 1 mol of electrons measured C/mol, $P_{\text{ion}}$ are the membrane permeabilities to the respective ions, and $[\text{ion}]_{\text{in}}$ and $[\text{ion}]{_\text{out}}$ are the intracellular and extracellular ion concentrations.

This equation accounts for the contributions of sodium, potassium, and chloride ions to the membrane potential. When calcium ions play a significant role, their terms can be included as well.

The permeabilities of $P_{\text{ion}}$ can be modulated by gating variables that depend on membrane potential and other factors. To model the time-dependent changes in ion concentrations and permeabilities, we consider:

\begin{align} \frac{d[\text{ion}]{\text{in}}}{dt} &= - \frac{I{\text{ion}}}{z_{\text{ion}} F V_{\text{cell}}} \ I_{\text{ion}} &= P_{\text{ion}} z_{\text{ion}}^2 F^2 \left( \frac{V}{RT} \right) \left( [\text{ion}]_{\text{in}} - [\text{ion}]_{\text{out}} e^{-z_{\text{ion}} F V / RT} \right) / \left( 1 - e^{-z_{\text{ion}} F V / RT} \right) \end{align}

where $I_{\text{ion}}$ is the ionic current, $z_{\text{ion}}$ is the valence of the ion, and $V_{\text{cell}}$ is the cell volume.

\subsubsection{Gating Variable Dynamics}

The permeabilities $P_{\text{ion}}$ can be modulated by gating variables. These parameters describe how ion channels in the cell membrane open or close in response to changes in the membrane potential ($V_mem$) and follow first-order kinetics:

\begin{equation} \frac{dP_{\text{ion}}}{dt} = \alpha_P(V)(\max(P_{ion}) - P_{\text{ion}}) - \beta_P(V) P_{\text{ion}} \end{equation}

Here, $\alpha_P(V)$ and $\beta_P(V)$ are voltage-dependent rate constants, and $P_{\text{ion,max}}$ is the maximum permeability for the ion.

\subsubsection{Calcium-Dependent signalling}

Calcium influx affects downstream signalling pathways, and its concentration is influenced by the membrane potential through voltage-gated calcium channels. The intracellular calcium concentration dynamics can be modeled as:

\begin{equation} \frac{d[\text{Ca}^{2+}]_{\text{in}}}{dt} = - \frac{I_{\text{Ca}}}{2 F V_{\text{cell}}} - k_{\text{pump}} [\text{Ca}^{2+}]_{\text{in}} \end{equation}

where $I_{\text{Ca}}$ is the calcium ionic current calculated similarly to $I_{\text{ion}}$, and $k_{\text{pump}}$ represents the rate of calcium removal from the cytosol via pumps.

The activation of calmodulin (CaM) and calcineurin proceeds as follows:

\begin{align} \frac{d[\text{Ca}^{2+}\text{-CaM}]}{dt} &= k_{\text{on}} [\text{Ca}^{2+}]_{\text{in}} [\text{CaM}] - k_{\text{off}} [\text{Ca}^{2+}\text{-CaM}] \\ \ \frac{d[\text{Calcineurin}_{active}]}{dt} &= k_{\text{on}} [\text{Ca}^{2+}\text{-CaM}] [\text{Calcineurin}] - k_{\text{off}} [\text{Calcineurin}_{active}] \end{align}

\subsection{Reinforcement Learning Framework}

\begin{figure}[H]
    \centering
    \includegraphics[width=0.8\textwidth]{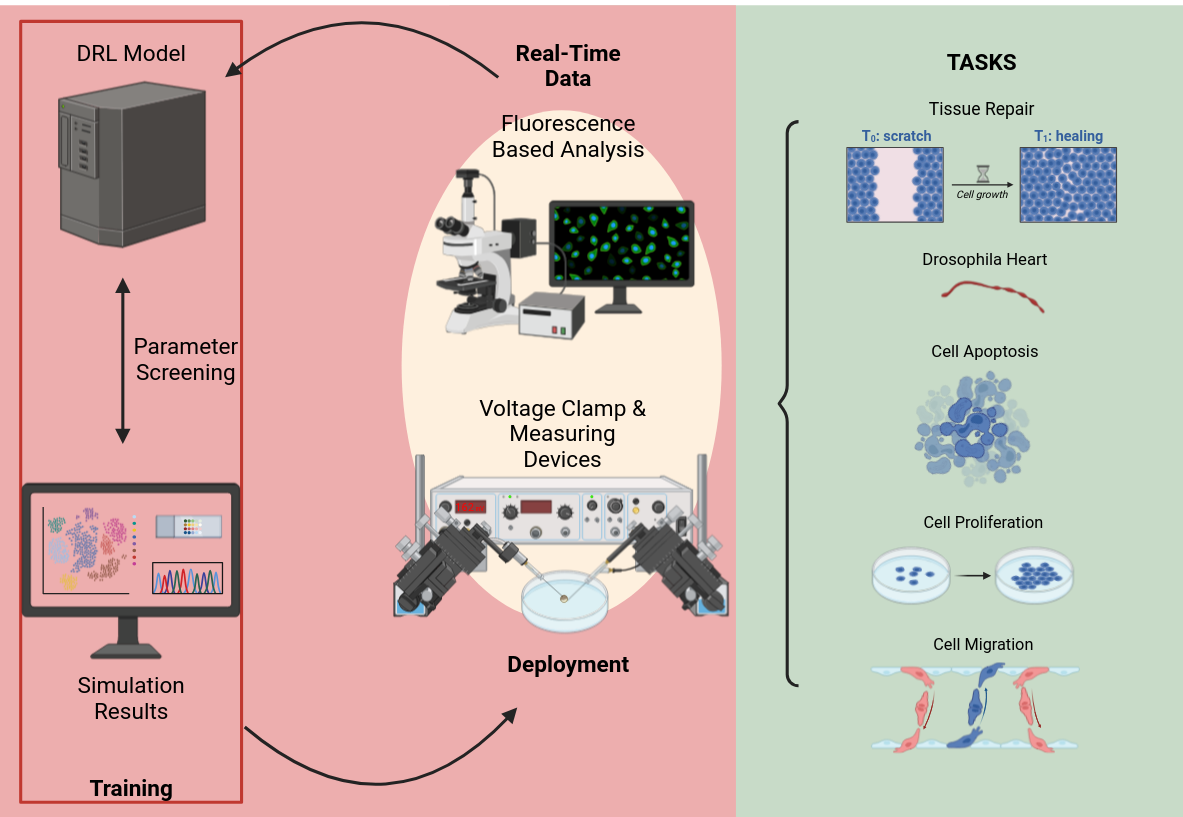} 
    \caption{Illustration of the DRL framework, from training to deployment (See Appendix \ref{sec:experiment} for a description of the experimental setup).}
    \label{fig:rl}
\end{figure}

DRL has achieved remarkable success in mastering complex games with very high dimensional state representations, as demonstrated by algorithms like AlphaGo \cite{Silver2016}, AlphaZero \cite{silver2017mastering}, MuZero \cite{Schrittwieser_2020}, OpenAI Five for \textit{Dota 2} \cite{openai2019dota}, and AlphaStar for \textit{StarCraft II} \cite{vinyals2019grandmaster}. These algorithms have showcased the ability of RL agents to learn optimal strategies in environments with vast state and action spaces, long time horizons, and partial observability \cite{deCarvalho2024deep}.

The proposed experimental setups that follow abide by a three-stage pipeline in which the algorithm first manipulates and makes predictions via a digital twin replica of the in vivo cells or tissues. This process allows for feedback loops before applying the outputs to the actual biological system (see Fig. \ref{fig:rl}).

The agent's state at time $t$, denoted $s_t$, includes variables such as ion concentrations ($\text{Na}^+$, $\text{K}^+$, $\text{Cl}^-$, $\text{Ca}^{2+}$, $\text{H}^+$), the membrane potential: $V$ calculated using the GHK equation, permeabilities: $P_{\text{Na}}$, $P_{\text{K}}$, $P_{\text{Cl}}$, modulated by gating variables, the gating variables, the protein activation states ($\text{Calmodulin}$, $\text{Calcineurin}_{active}$, $\text{Akt}$), pH levels, and concentrations of serotonin and butyrate.

The action space is defined by the agent's ability to set the membrane potential:

\begin{equation}
a_t = V_{\text{set}} \in [V_{\text{min}}, V_{\text{max}}]
\end{equation}

The agent's action influences the permeabilities $P_{\text{ion}}$ or gating variables that, in turn, affect the membrane potential.

The reward function $R_t$ should be constructed so as to encourage the agent to achieve homeostasis and desired activation levels of specific proteins, now with the membrane potential calculated via the GHK equation:

\begin{align} 
R_t &= w_{\text{homeostasis}} R_{\text{homeostasis}} + w_{\text{signalling}} R_{\text{signalling}} + w_{\text{efficiency}} R_{\text{efficiency}} \\
R_{\text{homeostasis}} &= - \sum_{\text{ions}} \left( \frac{[\text{ion}]_{\text{in}} - [\text{ion}]_{\text{desired}}}{\Delta [\text{ion}]} \right)^2 - \left( \frac{\text{pH}_{\text{in}} - \text{pH}_{\text{desired}}}{\Delta \text{pH}} \right)^2 \\ 
R_{\text{signalling}} &= - \left( \frac{[\text{Akt}]_{\text{desired}} - [\text{Akt}]_{\text{active}}}{\Delta [\text{Akt}]} \right)^2 - \left( \frac{[\text{Calcineurin}]_{\text{desired}} - [\text{Calcineurin}_{active}]}{\Delta [\text{Calcineurin}]} \right)^2 \\ 
R_{\text{efficiency}} &= - \left( \frac{\sum_{\text{ions}} |I_{\text{ion}}|}{I_{\text{max}}} \right)^2 
\end{align}

The agent adjusts the permeabilities $P_{\text{ion}}$ by influencing the gating variables through $V_{\text{set}}$. This action affects the membrane potential calculated by the GHK equation and, consequently, the ionic currents and downstream cellular processes.

To solve this problem, DRL algorithms suitable for continuous action spaces can be employed. Algorithms such as Deep Deterministic Policy Gradient (DDPG), Twin Delayed DDPG (TD3), and Soft Actor-Critic (SAC) utilize actor-critic architectures to learn optimal policies in continuous domains. These simpler models can be initially used as a proof of concept, before elaborating or attempting to deploy the DRL options mentioned above.

\subsubsection{Measurement of Variables and Real-Time Modelling Tools}

Implementing this framework requires measuring the necessary variables and modelling the system in real-time. Potential methods and tools include (See Appendix \ref{sec:experiment} for a description of the experimental setup):

\begin{itemize}
    \item \textbf{Optogenetics and Voltage-Sensitive Dyes}: These techniques could allow precise control and monitoring of membrane potentials in cells and tissues with high spatial and temporal resolution.
    \item \textbf{Fluorescent Reporters and Biosensors}: Gene expression levels, differentiation status, and other cellular properties might be tracked using fluorescent markers and biosensors.
    \item \textbf{Live Imaging Microscopy}: Techniques like two-photon microscopy could enable real-time observation of tissue morphology, cell migration, and topology changes.
    \item \textbf{Computational modelling Platforms}: Software such as COMSOL Multiphysics or custom-developed agent-based models could simulate bioelectric fields and cellular interactions, providing a virtual environment for training the DRL agent.
    \item \textbf{High-Performance Computing Resources}: Utilizing GPUs and distributed computing systems may be necessary to handle the computational demands of training complex DRL models.
    \item \textbf{Neuromorphic Computing}: Neuromorphic computing, inspired by biological neural architectures, could facilitate rapid and energy-efficient simulations of bioelectric activity in cellular systems. By simulating spiking neural networks (SNNs) and event-based processing, neuromorphic hardware can model complex cellular interactions in real time, possibly mimicking the dynamics of bioelectric signalling. Neuromorphic systems integrate memory and processing, allowing them to handle high-dimensional, continuous data from biological tissues while maintaining low power consumption and high parallelism. Neuromorphic systems could enable the DRL agent to interact with bioelectric simulations at high temporal resolutions, capturing dynamic feedback loops important in precise bioelectric control and tissue modulation (See Appendix \ref{sec:neuromorphic}).
\end{itemize}

\subsubsection{Current Methods and the Advantage of Using DRL}

Currently, controlling bioelectric signals in tissue engineering and regenerative medicine relies heavily on predefined protocols and heuristic methods. These approaches often lack the flexibility and adaptability required to handle the complex, dynamic interactions within biological systems. Specifically, existing methods use static sequences of bioelectric interventions that do not account for real-time feedback or the evolving state of the tissue. Additionally, these methods depend on expert knowledge to set bioelectric parameters, which can be time-consuming and may not capture the optimal intervention strategies. Traditional approaches also struggle to adapt to unforeseen changes or variations in tissue responses, leading to suboptimal recovery outcomes.

In contrast, using DRL offers several significant advantages. DRL agents learn optimal policies through interaction with the environment, allowing them to adapt to real-time feedback and dynamic changes in tissue states. By maximizing a cumulative reward function, DRL can discover intervention sequences that more effectively guide tissue regeneration and recovery. Moreover, DRL methods can handle high-dimensional state and action spaces, making them suitable for complex biological systems with numerous interacting variables. Finally, DRL reduces the reliance on manual tuning and expert intervention, streamlining the process of bioelectric signal control and enhancing overall efficiency.

\subsubsection{Modelling Heart Topology and Bioelectric Patterns}

The topology of the \textit{Drosophila} heart refers to the spatial arrangement and connectivity of its structural components, including cells, tissues, and the overall geometry of the organ. To accurately model the heart's assembly, it is important to define and quantify its topology and bioelectric patterns mathematically.

The heart's topology can be represented using computational geometry and graph theory. The spatial distribution of cells can be modelled using meshes or point clouds, while the connectivity between cells can be represented as graphs where nodes correspond to cells and edges represent physical or functional connections.

To quantify differences between the current and desired heart topology, metrics such as the Hausdorff distance can be employed:

\begin{equation}
d_H(A, B) = \max\left\{\sup_{a \in A} \inf_{b \in B} \|a - b\|, \sup_{b \in B} \inf_{a \in A} \|a - b\|\right\}
\end{equation}

where $A$ and $B$ are point sets representing the model heart and the real heart, respectively. And $sup$ refers to the supremum, which determines the largest "worst-case" distance between the sets $A$ and $B$.

Similarly, bioelectric patterns, which encompass the distribution of membrane potentials across the heart tissue, can be compared using spatial integrals of the squared differences:

\begin{equation}
\text{Bioelectric Error} = \int_{\text{Heart Volume}} \left| V_{\text{mem}}^{\text{model}}(\mathbf{r}) - V_{\text{mem}}^{\text{target}}(\mathbf{r}) \right|^2 dV
\end{equation}

where $V_{\text{mem}}^{\text{model}}(\mathbf{r})$ is the membrane potential at position $\mathbf{r}$ in the model, and $V_{\text{mem}}^{\text{target}}(\mathbf{r})$ is the desired membrane potential at that position.

By incorporating these measures into the model, the agent can attempt to evaluate the difference between the current bioelectric patterns and the target bioelectric patterns, as well as the differences in topology, guiding the assembly process toward the desired outcome.

In bioelectric experiments conducted by Michael Levin’s team, similar principles are applied to study pattern formation in regenerating tissues, such as frog embryos. Levin’s group uses voltage-sensitive dyes, such as DiBAC4(3) and CC2-DMPE, to detect membrane potential differences across cells. These dyes change their fluorescence based on the membrane potential, providing real-time spatial and temporal mapping of the bioelectric state in both normal and abnormal tissues. Through fluorescent microscopy, these bioelectric states are visualized and quantified, allowing the identification of deviations in abnormal tissues compared to healthy ones \cite{adams2013endogenous}.

To induce a normal bioelectdrcfx   ric pattern in an abnormal frog embryo, ion channels are pharmacologically modulated or optogenetics is used. Optogenetics involves light-sensitive ion channels introduced into the tissue, normally through CRISP-R, enabling precise control over membrane potential with light. By manipulating the ion flow and thus the membrane potential, researchers have been able to correct the bioelectric patterns in the abnormal tissue to match those of the healthy, normal tissue. For instance, an abnormal head pattern in a frog can be corrected by restoring the normal bioelectric gradient, guiding the cells to form a correctly structured and functional head \cite{adams2013endogenous}.

This approach, known as bioelectric modulation, effectively uses the bioelectric pattern as a template for normal tissue development and regeneration. It demonstrates that bioelectric states serve as key regulators of tissue and organ formation, enabling control over developmental processes through targeted modulation of membrane potentials \cite{adams2013endogenous}.

The principles observed in previous bioelectric experiments are directly applicable to the assembly and morphogenesis of the \textit{Drosophila} heart. Just as bioelectric patterns can guide the formation of a frog's head by instructing cells through ion channel modulation, the developing heart in \textit{Drosophila} relies on bioelectric signals to ensure the correct arrangement and function of its tissues. By understanding and manipulating these bioelectric cues, we can attempt to control and make cells in the heart adopt the correct positions and connectivity, driving the proper morphological outcome. Thus, by incorporating bioelectric modulation into the model of heart assembly, it becomes possible to precisely control the morphogenesis of the heart in a similar way to how frog tissues are reprogrammed in the experiments mentioned above.

\subsection{Reinforcement Learning Framework for Physiological Recovery of a Damaged \textit{Drosophila Melanogaster} Heart}

In this expanded model, we propose that the agent's action space may involve global or spatially targeted control of membrane potential:

\begin{equation}
a_t = V_{\text{set}} \quad \text{or} \quad a_t: \mathbf{r} \rightarrow V_{\text{set}}(\mathbf{r})
\end{equation}

where:
\begin{itemize}
    \item $a_t$ is the action taken by the agent at time $t$.
    \item $V_{\text{set}}$ represents the membrane potential value set by the agent.
    \item $\mathbf{r}$ denotes spatial coordinates within the tissue.
    \item $V_{\text{set}}(\mathbf{r})$ is a function defining the membrane potential to be set at position $\mathbf{r}$.
\end{itemize}

The state at time $t$, denoted as $s_t$, includes detailed information about the cellular states, tissue morphology, bioelectric patterns, and heart topology:

\begin{equation}
s_t = (\mathbf{C}_t, \mathbf{M}_t, \mathbf{V}_t, \mathbf{T}_t) \in \mathcal{S}
\end{equation}

where:
\begin{itemize}
    \item $\mathbf{C}_t$ represents the set of cellular properties at time $t$.
    \item $\mathbf{M}_t$ denotes the tissue morphology at time $t$.
    \item $\mathbf{V}_t$ is the bioelectric pattern at time $t$.
    \item $\mathbf{T}_t$ represents the heart topology at time $t$.
    \item $\mathcal{S}$ is the state space.
\end{itemize}

\subsubsection*{Game Interpretation}

In this framework, we interpret the physiological recovery task as a strategic game. The objective of the game is to restore a damaged heart to its functional state by optimally manipulating bioelectric signals.

\subsubsection{State Space}

The state space $\mathcal{S}$ comprises all possible configurations of the heart tissue, encapsulating:

\begin{itemize}
    \item \textbf{Cellular States} ($\mathbf{C}_t$): A set of cellular properties at time $t$, where each cell $i$ has attributes such as:
    \begin{itemize}
        \item Membrane potential $V_{\text{mem},i}$.
        \item Gene expression levels $G_i$.
        \item Differentiation status $D_i$.
        \item Position $\mathbf{p}_i$.
    \end{itemize}
    Thus,
    \[
    \mathbf{C}_t = \{ (V_{\text{mem},i}, G_i, D_i, \mathbf{p}_i) \}_{i=1}^{N_t}
    \]
    where $N_t$ is the total number of cells at time $t$.
    \item \textbf{Tissue Morphology} ($\mathbf{M}_t$): The geometrical arrangement and structural properties of the tissue at time $t$.
    \item \textbf{Bioelectric Patterns} ($\mathbf{V}_t$): The spatial distribution of membrane potentials across the tissue at time $t$.
    \item \textbf{Heart Topology} ($\mathbf{T}_t$): The connectivity and organization of cells forming the heart structure at time $t$.
\end{itemize}

\subsubsection{Action Space}

The action space \( \mathcal{A} \) consists of all possible interventions the agent can apply to influence the bioelectric state of the tissue. In our experimental setup, the agent outputs a voltage mesh applied via the 3D microelectrode array, enabling spatially targeted modifications of the membrane potential across the tissue (See Appendix \ref{sec:experiment}).

Therefore, the action at time \( t \), denoted as \( a_t \), is a mapping from spatial coordinates to voltage values, forming a voltage mesh:

\begin{equation}
a_t: \mathbf{r} \mapsto V_{\text{set}}(\mathbf{r}), \quad \mathbf{r} \in \Omega
\end{equation}

where:
\begin{itemize}
    \item \( \mathbf{r} \) represents the spatial coordinates within the tissue domain \( \Omega \).
    \item \( V_{\text{set}}(\mathbf{r}) \) is the voltage value set by the agent at position \( \mathbf{r} \).
    \item \( \Omega \) is the spatial domain of the tissue covered by the microelectrode array.
\end{itemize}

Thus, the action space can be expressed as:

\begin{equation}
a_t \in \mathcal{A} = \left\{ V_{\text{set}}(\mathbf{r}) \mid \mathbf{r} \in \Omega \right\}
\end{equation}

\subsubsection{Goal}

The goal of the agent is to discover an optimal policy $\pi(a_t \mid s_t)$ that maximizes the expected cumulative reward, effectively guiding the tissue to reassemble into a functional heart. Formally, the objective is:

\begin{equation}
\max_{\pi} \, \mathbb{E}_{\pi} \left[ \sum_{t=0}^{T} \gamma^t R_t \right]
\end{equation}

where:
\begin{itemize}
    \item $\mathbb{E}_{\pi}$ denotes the expected value over trajectories following policy $\pi$.
    \item $\gamma \in [0,1)$ is the discount factor representing the importance of future rewards.
    \item $R_t$ is the reward received at time $t$.
    \item $T$ is the time horizon or total number of time steps.
\end{itemize}

\subsubsection{Game Dynamics}

At each time step $t$, the agent observes the current state $s_t$ and selects an action $a_t$ according to its policy $\pi$. The environment (the biological system) then transitions to a new state $s_{t+1}$ based on the dynamics of cellular processes influenced by the action $a_t$. The agent receives a reward $R_t$ reflecting the immediate impact of its action on achieving the desired heart recovery.

\newpage
\subsubsection{Reward Function}
The reward function is designed to guide the agent toward assembling a functional heart. It includes terms that reflect the desired outcomes in morphology, topology, bioelectric patterns, differentiation, proliferation, apoptosis, gene expression, migration, and efficiency:

\begin{align}
R_t &= R_{\text{Morph}} + R_{\text{Topo}} + R_{\text{Bioelec}} + R_{\text{Diff}} + 
R_{\text{Prolif}} + R_{\text{Apop}} + R_{\text{Gene}} + R_{\text{Mig}} + R_{\text{Eff}} \\
R_{\text{Morph}} &= -w_{\text{Morph}} \times \text{Morphology Error} \\
R_{\text{Topo}} &= -w_{\text{Topo}} \times \text{Topology Error} \\
R_{\text{Bioelec}} &= -w_{\text{Bioelec}} \times \text{Bioelectric Error} \\
R_{\text{Diff}} &= w_{\text{Diff}} \times \frac{N_{\text{Diff}}}{N_T} \\
R_{\text{Prolif}} &= w_{\text{Prolif}} \times \left( \frac{N_T - |N_T - N_{\text{Target}}|}{N_T} \right) \\
R_{\text{Apop}} &= -w_{\text{Apop}} \times \left( \frac{N_{\text{Apop}}}{N_T} \right) \\
R_{\text{Gene}} &= -w_{\text{Gene}} \sum_{i} \left( \frac{[\text{Gene}_i]_{\text{desired}} - [\text{Gene}_i]}{\Delta [\text{Gene}_i]} \right)^2 \\
R_{\text{Mig}} &= -w_{\text{Mig}} \times \frac{\text{Total Migration Error}}{\text{Max Error}} \\
R_{\text{Eff}} &= -w_{\text{Eff}} \times \left( \frac{\sum_{\text{cells}} I_{T_i}}{I_{\text{max}}} \right)
\end{align}

Above is a list of possible compounded ways of computing reward \footnote{
    $R_t$ is the total reward at time $t$.
    $R_{\text{Morph}}$ penalizes deviations from the desired tissue morphology.
    $\text{Morphology Error}$ quantifies the difference between current and target morphology.
    $w_{\text{Morph}}$ is the weight assigned to morphology error.
    $R_{\text{Topo}}$ penalizes deviations from the desired heart topology.
    $\text{Topology Error}$ quantifies the structural differences.
    $w_{\text{Topo}}$ is the weight for topology error.
    $R_{\text{Bioelec}}$ penalizes deviations from desired bioelectric patterns.
    $\text{Bioelectric Error}$ measures the difference in bioelectric states.
    $w_{\text{Bioelec}}$ is the weight for bioelectric error.
    $R_{\text{Diff}}$ rewards the proportion of correctly differentiated cells.
    $N_{\text{Diff}}$ is the number of correctly differentiated cells.
    $N_T$ is the total number of cells.
    $w_{\text{Diff}}$ is the weight for differentiation.
    $R_{\text{Prolif}}$ rewards appropriate cell proliferation.
    $N_{\text{Target}}$ is the target number of cells.
    $w_{\text{Prolif}}$ is the weight for proliferation.
    $R_{\text{Apop}}$ penalizes unnecessary apoptosis.
    $N_{\text{Apop}}$ is the number of apoptotic cells.
    $w_{\text{Apop}}$ is the weight for apoptosis.
    $R_{\text{Gene}}$ penalizes deviations in gene expression levels.
    $[\text{Gene}_i]$ is the expression level of gene $i$.
    $[\text{Gene}_i]_{\text{desired}}$ is the desired expression level.
    $\Delta [\text{Gene}_i]$ is the acceptable variation in expression.
    $w_{\text{Gene}}$ is the weight for gene expression.
    $R_{\text{Mig}}$ penalizes incorrect cell migration.
    $\text{Total Migration Error}$ quantifies deviations in cell positions.
    $\text{Max Error}$ is the maximum possible migration error.
    $w_{\text{Mig}}$ is the weight for migration.
    $R_{\text{Eff}}$ penalizes energy inefficiency.
    $I_{T_i}$ is the total ionic current for cell $i$.
    $I_{\text{max}}$ is the maximum allowable total ionic current.
$w_{\text{Eff}}$ is the weight for efficiency.
}.

\subsubsection{Output of the DRL and Its Application to Cell Control}

The primary output of the DRL agent is the optimal policy $\pi(a_t \mid s_t)$, which dictates the sequence of actions $a_t$ (i.e., the specific membrane potential settings $V_{\text{set}}$) to be applied at each state $s_t$. This policy serves as a decision-making framework that guides the agent in manipulating the bioelectric signals across the heart tissue. Specifically, the DRL agent uses the learned policy to control membrane potentials either globally or at specific spatial locations, thereby influencing cellular behaviours such as differentiation, proliferation, and apoptosis. These targeted interventions guide the organization and structural formation of heart tissue, ensuring proper topology and functional integrity. Through precise control of bioelectric signals, the agent facilitates efficient cellular processes that contribute to the overall recovery and functionality of the heart.

\section{Causal Inference in Deep Reinforcement Learning Framework}
Understanding the causal relationships between bioelectric signals and cellular behaviours could aid control and manipulation of morphogenetic processes. Integrating Judea Pearl’s do-calculus into our DRL framework could allow for more precise causal inference, helping with the agent's decision-making processes \cite{pearl_book, Pearl2010, Pearl1995}. This graphical representation can also function as an interpretability and explainability tool, regardless of its impact in predicting desirable actions for specific outcomes. In the latter case, we simply map the agent's reactions to the status of the in vivo cells and environmental factors while recording the outcome of these actions. Over time, we will approximate probability distributions of the action space and the actual outcomes of these actions, facilitating analysis over the experimental data \cite{pearlcauses} (See Appendix \ref{sec:docalc}).

\subsection{Causal Graphs for Bioelectric Signalling - Examples}
Causal graphs, represented as Directed Acyclic Graphs (DAGs), visually depict the causal relationships between variables. In the context of bioelectric signalling, such a representation may begin with scarce variables, including:

\begin{itemize}
    \item \textbf{Bioelectric Signals ($V_{\text{mem}}$)}: Voltage gradients across cell membranes.
    \item \textbf{Cellular behaviours}: Proliferation, differentiation, migration, apoptosis, and gene expression.
    \item \textbf{DRL Agent Actions ($A_t$)}: Interventions to manipulate $V_{\text{mem}}$.
    \item \textbf{Environmental Factors ($E$)}: External stimuli and conditions affecting cellular responses.
\end{itemize}

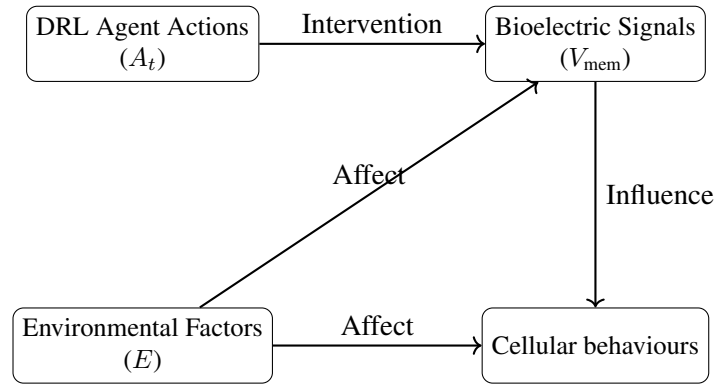
\begin{figure}[H]
    \centering
    \begin{tikzpicture}[
        node distance=3cm and 3cm,
        main node/.style={
            draw,
            rectangle,
            rounded corners,
            align=center,
            minimum width=2.5cm,
            minimum height=1cm,
            font=\small
        },
        arrow/.style={
            ->,
            thick
        }
    ]
        \node[main node] (Actions) {DRL Agent Actions\\($A_t$)};
        \node[main node] (Vmem) [right=of Actions] {Bioelectric Signals\\($V_{\text{mem}}$)};
        \node[main node] (behaviours) [below=of Vmem] {Cellular behaviours};
        \node[main node] (Environment) [below=of Actions] {Environmental Factors\\($E$)};
        
        \draw[arrow] (Actions) -- (Vmem) node[midway, above] {Intervention};
        \draw[arrow] (Vmem) -- (behaviours) node[midway, right] {Influence};
        \draw[arrow] (Environment) -- (Vmem) node[midway, above] {Affect};
        \draw[arrow] (Environment) -- (behaviours) node[midway, above] {Affect};
    \end{tikzpicture}
    \caption{Causal Graph Representing the Relationships Between DRL Actions, Bioelectric Signals, Cellular behaviours, and Environmental Factors.}
    \label{fig:causal_graph}
\end{figure}

\begin{figure}[H]
    \centering
    \begin{tikzpicture}[
        node distance=3cm and 3cm,
        main node/.style={
            draw,
            ellipse,
            align=center,
            minimum width=2.5cm,
            minimum height=1cm,
            font=\small
        },
        arrow/.style={
            ->,
            thick
        }
    ]
        \node[main node] (Actions) {DRL Agent Actions\\($A_t$)};
        \node[main node] (Vmem) [right=of Actions] {Bioelectric Signals\\($V_{\text{mem}}$)};
        \node[main node] (Calcium) [below=of Vmem] {Intracellular Ca$^{2+}$};
        \node[main node] (GeneExpr) [right=of Calcium] {Gene Expression};
        \node[main node] (behaviours) [below=of Calcium] {Cellular behaviours};
        \node[main node] (Environment) [below=of Actions] {Environmental Factors\\($E$)};
        
        \draw[arrow] (Actions) -- (Vmem) node[midway, above] {Intervention};
        \draw[arrow] (Vmem) -- (Calcium) node[midway, left] {Influence};
        \draw[arrow] (Calcium) -- (GeneExpr) node[midway, above] {Regulation};
        \draw[arrow] (GeneExpr) -- (behaviours) node[midway, above] {Control};
        \draw[arrow] (Environment) -- (Vmem) node[midway, below] {Affect};
        \draw[arrow] (Environment) -- (behaviours) node[midway, right] {Affect};
    \end{tikzpicture}
    \caption{Expanded Causal Graph Including Calcium signalling and Gene Expression.}
    \label{fig:expanded_causal_graph}
\end{figure}
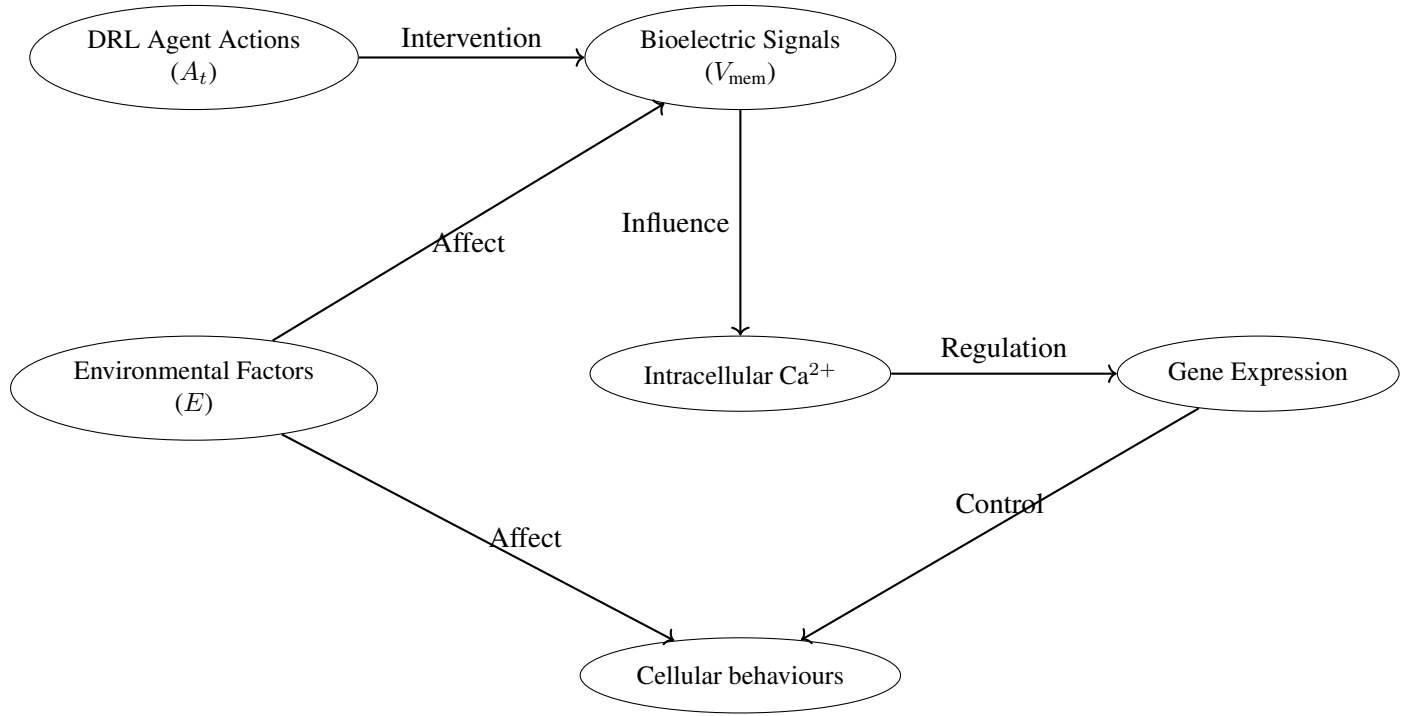

\section{Expected Outcomes}

This research aims to address several of the critical questions outlined in the introduction by developing and implementing a DRL framework for real-time control of bioelectric signals. The expected outcomes are as follows:

\begin{enumerate} \item \textbf{Predicting and Manipulating $V_{\text{mem}}$ Distributions (Questions 1, 12, 14):} By integrating the DRL framework with advanced optogenetic tools and voltage reporters, we expect to develop methods for predicting and controlling the spatial and temporal distributions of membrane potentials ($V_{\text{mem}}$) within cell groups. This will address the challenge of accurately mapping and manipulating $V_{\text{mem}}$ values across tissues in real-time, enabling precise bioelectric modulation.

\item \textbf{Understanding Bioelectric Regulation of Morphogenesis (Questions 2, 3, 10):} The DRL agent will autonomously discover bioelectric manipulation strategies that lead to desired morphological outcomes, effectively unraveling how bioelectric signals regulate anatomical growth and tissue formation. This will enhance our ability to identify and manipulate the morphological blueprints of complex organs without genetic intervention.

\item \textbf{Elucidating Bioelectric Mechanisms of Organ Cohesion (Questions 4, 9):} By modelling and controlling bioelectric signals in tissues and organs, we expect to uncover the mechanisms by which collections of cells coordinate their bioelectric states to function cohesively as an organ. This includes understanding how cells compare and synchronize their bioelectric states across distances to maintain homeostasis and structural stability.

\item \textbf{Exploring Cellular Decision-Making Capacities (Questions 5, 13):} The computational models developed will simulate cells as agents with decision-making capabilities influenced by bioelectric signals. This will shed light on the degrees of freedom cells possess and whether bioelectric circuits in non-neural cells can store and process information akin to neural circuits.

\item \textbf{Preventing Cellular Defection and Cancer (Questions 6, 7, 11):} Through artificial manipulation of bioelectric signalling, we aim to investigate strategies to prevent cellular defection, such as cancerous transformations, by enforcing conformity within tissues. The development of new synthetic biology tools for top-down programming of bioelectric circuits could have significant applications in regenerative medicine and cancer therapy.

\item \textbf{Optimizing Experimental Models (Question 8):} By utilizing DRL and computational modelling, we will determine the minimal number of cells required to effectively study bioelectric phenomena, optimizing experimental designs and resource utilization.

\item \textbf{Developing Quantitative Bioelectric Models (Question 10):} The data collected will contribute to the development of quantitative models of bioelectric circuits that reliably store stable patterning information during morphogenesis. These models will improve our predictive capabilities and understanding of bioelectric-driven development.

\item \textbf{Advancing Bioelectric Manipulation Techniques (Questions 12, 14):} The research will contribute to expanding optogenetic techniques and developing new voltage reporters for better in vivo modulation and observation of bioelectric states in real-time, enhancing our ability to manipulate bioelectric signals in large, nonexcitable cell groups.
\end{enumerate}

By achieving these outcomes, this research will significantly advance our understanding of bioelectricity in developmental biology and provide innovative tools for manipulating bioelectric signals to control tissue growth and regeneration.

\newpage

\section{Appendix}

\subsection{Experimental Setup}\label{sec:experiment}
To validate the proposed Deep Reinforcement Learning (DRL) framework for real-time control of bioelectric signals in organ-level regeneration, we design an experimental setup involving the adult heart organ of \textit{Drosophila melanogaster}. This setup includes inducing a controlled heart injury without completely killing the heart, allowing us to test the DRL agent's ability to modulate bioelectric signals to promote physiological recovery.

\subsubsection{Induction of Heart Defects}
We employ genetic techniques to induce heart-specific defects in larvae that manifest in the adult heart. Using the GAL4/UAS system, we express genes that disrupt heart function specifically in cardiac tissues. For example:
\begin{itemize}
    \item \textbf{Expression of Pro-Apoptotic Genes}: By driving the expression of genes such as \textit{reaper} or \textit{hid} under the control of a heart-specific promoter like \textit{tinC-GAL4}, we induce apoptosis in a subset of cardiac cells.
    \item \textbf{RNA Interference (RNAi)}: Using UAS-RNAi lines targeting essential cardiac genes (e.g., \textit{hand}, \textit{dMEF2}), we knock down gene expression specifically in the heart, leading to structural or functional defects \cite{Lesch2010}.
\end{itemize}

These genetic manipulations are performed during the larval stages, allowing the defects to develop as the larvae mature into adults. The adult flies with heart defects are then used for the experiments.

Alternatively, we can induce mechanical injury to the adult heart without causing lethality:

\begin{itemize}
    \item \textbf{Laser Ablation}: Focused laser pulses are used to create precise injuries in the heart tissue of anesthetized adult flies. This method allows for controlled damage to specific regions of the heart.
\end{itemize}

\subsubsection{Isolation and Maintenance of the Adult Heart Organ}

The adult flies with induced heart defects are anesthetized on ice and dissected to isolate the heart organ (dorsal vessel). The dissection is performed in oxygenated artificial hemolymph solution containing necessary ions and nutrients to mimic the in vivo environment.

The isolated heart organ is transferred to a specialized perfusion chamber designed to maintain tissue viability. The chamber provides continuous flow of oxygenated artificial hemolymph at 25°C. The heart is pinned gently to a Sylgard-coated dish to stabilize it without causing additional damage \cite{Wasserthal2007}.

A custom-designed three-dimensional (3D) microelectrode array (MEA) is fabricated to interface directly with the ex vivo heart organ. The MEA consists of microelectrodes arranged to match the geometry of the \textit{Drosophila} heart, enabling localized application of electrical stimuli and recording of bioelectric signals \cite{scalableElectro}.

The heart is carefully positioned onto the MEA so that the microelectrodes make contact with the myocardial surface. To improve electrical coupling, the MEA surface is coated with conductive polymers. The integration is performed under a stereomicroscope to ensure precise alignment and minimal stress on the tissue.

The DRL agent outputs a voltage mesh corresponding to the spatial distribution of electrical stimuli to be applied across the heart organ. A custom software interface translates the DRL outputs into stimulation commands for the MEA, controlling parameters such as voltage amplitude, frequency, and waveform for each electrode.

A multi-channel stimulator delivers the specified electrical stimuli to the heart via the MEA. The stimulator is synchronized with the DRL agent, allowing real-time adjustments based on the heart's responses.

Bioelectric signals from the heart, such as action potentials and conduction velocities, are recorded using the MEA. The signals are amplified, filtered, and digitized for analysis. Optical mapping techniques using voltage-sensitive dyes (e.g., Di-4-ANEPPS) are employed to visualize electrical activity across the heart surface.

The recorded data are fed back to the DRL agent as part of the state $s_t$, enabling the agent to update its policy based on the observed cardiac responses. This closed-loop system allows the DRL agent to learn and adapt in real time.

This same data is collected from healthy heart samples so as to compute the reward functions later on.

\subsection{Decoding the Bioelectric Language}

\subsubsection{Current Understanding of Bioelectric Signals}

Bioelectric signals are fundamental to the regulation of cellular and developmental processes. They are generated by the movement of ions across cellular membranes, creating voltage gradients ($V_{\text{mem}}$) that influence cell behaviour. Recent research has begun to map the bioelectric patterns associated with various cellular states and developmental processes, revealing that these patterns play a crucial role in tissue formation, regeneration, and organogenesis \cite{Levin2009, Levin2014wh, adams2013endogenous}.

Just as the genome has been sequenced to understand genetic contributions to biology, efforts have been made to map bioelectric patterns in organisms. Techniques such as voltage-sensitive dyes and fluorescent reporters have allowed visualization of membrane potential distributions in living tissues. Studies have shown that specific bioelectric patterns correlate with particular developmental outcomes. For example, the formation of the anterior-posterior axis in \textit{Xenopus laevis} embryos is regulated by bioelectric cues \cite{adams2013endogenous, Pai2012}.

However, unlike the genome, the bioelectric "map" is dynamic and context-dependent. The spatial and temporal variability of bioelectric signals across different cell types and developmental stages makes comprehensive mapping challenging. Current maps are incomplete and often organism-specific, limiting the generalization of findings \cite{McCaig2005, Sundelacruz2009}.

Cells interpret bioelectric signals through voltage-gated ion channels, transporters, and voltage-sensitive signalling pathways. Changes in membrane potential can influence gene expression, cell proliferation, differentiation, migration, and apoptosis. For instance, depolarization of membrane potential has been shown to promote stem cell proliferation, while hyperpolarization can induce differentiation \cite{Levin2009, Levin2014wh, Blackiston2011}.

Despite these insights, the precise mechanisms by which cells transduce bioelectric signals into specific biochemical responses remain poorly understood. The complexity arises from the interplay between electrical, chemical, and mechanical signals within the cellular microenvironment. Additionally, the same bioelectric signal can elicit different responses depending on the cell type and its developmental context \cite{Levin2014wh}.

\subsubsection{Standardizing Bioelectric Codes}
The concept of a "bioelectric code" refers to the idea that specific patterns of membrane potential and ion fluxes can be correlated with distinct physiological outcomes. Efforts to standardize this code involve cataloging the bioelectric signatures associated with various cellular functions and developmental processes.

Current progress includes identifying bioelectric states that correspond to particular tissue types or regenerative capabilities. For example, specific voltage gradients have been linked to limb regeneration in amphibians.


Our proposed DRL framework offers a novel approach to decoding the bioelectric language by enabling real-time control and analysis of bioelectric signals in biological systems.

By employing the DRL agent to manipulate bioelectric signals and observe the resulting cellular responses, we can iteratively build a comprehensive map of bioelectric patterns and their associated outcomes. The agent's ability to explore a vast action space and adapt based on feedback allows for the discovery of previously unknown bioelectric configurations that lead to desired cellular behaviours.


The DRL framework can help uncover how cells interpret bioelectric signals by identifying the causal relationships between specific membrane potential manipulations and cellular responses. By integrating causal inference techniques, as discussed in our framework, we can isolate the effects of bioelectric changes on gene expression, signal transduction pathways, and phenotypic outcomes.


Through systematic experimentation and data collection facilitated by the DRL agent, we can begin to standardize the bioelectric code. The agent's policy, which maps states to optimal actions, effectively represents a functional mapping between bioelectric patterns and physiological outcomes. By analyzing this policy, we can extract generalizable rules and bioelectric signatures associated with specific cellular functions.


\subsubsection{Advantages Over Traditional Methods}

Traditional approaches to decoding the bioelectric language rely on manual experimentation and are limited by the complexity and high dimensionality of biological systems. Our DRL framework automates the exploration of bioelectric space, efficiently handling the complexity through advanced algorithms inspired by game-solving AI.


Moreover, the integration of real-time data acquisition techniques, such as optogenetics and voltage-sensitive imaging, allows the agent to adjust its actions based on immediate feedback, accelerating the discovery process.


\subsection{Introduction to Neuromorphic Computing}
Neuromorphic computing is an emerging field that seeks to emulate the architecture and functioning of the biological nervous system in hardware and software systems \cite{Indiveri2011}. These systems typically utilize spiking neural networks (SNNs) and other brain-inspired architectures to achieve highly parallel and energy-efficient processing. Unlike traditional von Neumann architectures, neuromorphic systems integrate memory and computation, potentially offering advantages in real-time data processing and low-latency responses \cite{Benjamin2014}.

\subsubsection{Analogies Between Neuromorphic Systems and Cellular Bioelectric signalling}
Biological cells communicate through bioelectric signals, primarily mediated by the flow of ions across cell membranes, which generate voltage gradients ($V_{\text{mem}}$) essential for various cellular functions \cite{Levin2014wh}. Similarly, neuromorphic systems employ spiking neurons that communicate via discrete electrical pulses, akin to action potentials in biological neurons \cite{Maass1997}. This parallel suggests a potential framework where neuromorphic hardware could model and interact with cellular bioelectric behaviours, enabling more biologically plausible simulations and controls within the DRL framework.

While the application of neuromorphic computing to biological systems is still in its nascent stages, preliminary studies indicate promising directions. For instance, \cite{Roy2019} explored the use of SNNs to model neural dynamics in biological tissues, achieving real-time simulations of neuronal networks.

\subsubsection{Hypothetical Integration of Neuromorphic Computing with the DRL Framework}\label{sec:neuromorphic}
Integrating neuromorphic computing into the DRL framework could offer several speculative advantages for controlling bioelectric signals in real-time:

\begin{enumerate}
    \item \textbf{Enhanced Real-Time Processing}: Neuromorphic hardware's ability to process information with minimal latency may allow the DRL agent to interact more effectively with dynamic biological systems.
    \item \textbf{Energy Efficiency}: The low power consumption characteristic of neuromorphic systems could facilitate large-scale simulations of cellular networks without significant energy overhead.
    \item \textbf{Parallelism}: The inherently parallel nature of neuromorphic architectures aligns with the distributed communication patterns observed in biological tissues, potentially enabling more accurate and scalable models.
\end{enumerate}

To conceptualize the integration of neuromorphic computing within the DRL framework, we propose the following tentative mathematical model:

\begin{equation}
\mathcal{N}: \left\{
    \begin{array}{ll}
        \mathbf{V}_{\text{mem}}(t) &= \mathcal{F}_{\text{neuromorphic}}(\mathbf{I}(t), \mathbf{V}_{\text{mem}}(t-1)) \\
        \mathbf{S}_t &= \mathcal{G}(\mathbf{V}_{\text{mem}}(t), \mathbf{X}_t) \\
        \mathbf{A}_t &= \pi_{\theta}(\mathbf{S}_t) \\
        \mathbf{V}_{\text{mem}}(t+1) &= \mathcal{N}(\mathbf{A}_t, \mathbf{V}_{\text{mem}}(t))
    \end{array}
\right.
\end{equation}

where:
\begin{itemize}
    \item $\mathcal{N}$ denotes the neuromorphic processor.
    \item $\mathbf{V}_{\text{mem}}(t)$ is the vector of membrane potentials at time $t$.
    \item $\mathcal{F}_{\text{neuromorphic}}$ represents the neuromorphic model simulating bioelectric dynamics.
    \item $\mathbf{I}(t)$ is the input current vector influencing membrane potentials.
    \item $\mathbf{S}_t$ is the state observed by the DRL agent at time $t$.
    \item $\mathcal{G}$ is the function mapping membrane potentials and additional variables $\mathbf{X}_t$ to the agent's state.
    \item $\mathbf{A}_t$ is the action taken by the agent at time $t$, determined by the policy $\pi_{\theta}$.
\end{itemize}

This framework is speculative and aims to provide a foundation for future exploration into how neuromorphic computing might enhance the DRL approach for bioelectric signal manipulation.

\subsubsection{Training and Deployment Strategy}
The training and deployment strategy for integrating neuromorphic computing with a DRL agent begins with initializing the neuromorphic model. This initial setup establishes baseline bioelectric states reflective of the biological system being studied, providing a foundation for simulating bioelectric dynamics. At each timestep \( t \), the system observes and extracts the current state \( \mathbf{S}_t \) from the neuromorphic simulation, capturing essential bioelectric parameters necessary for accurate modelling.

Based on the observed state \( \mathbf{S}_t \), the DRL agent then selects an action \( \mathbf{A}_t \) according to its policy \( \pi_{\theta} \). This chosen action is applied to the neuromorphic processor to modulate membrane potentials, directly influencing bioelectric activity within the simulation. Following this modulation, a reward \( R_t \) is calculated by assessing how closely the resulting membrane potential \( \mathbf{V}_{\text{mem}}(t+1) \) aligns with desired bioelectric patterns or other biological targets. This reward feedback is then used to update the agent’s policy \( \pi_{\theta} \), enabling the agent to refine its decision-making capabilities and improve over time.

\subsubsection{Potential Benefits and Challenges}
The integration of neuromorphic computing within this DRL framework offers several potential benefits. Neuromorphic systems are notably scalable, allowing them to handle larger, more complex biological simulations efficiently. Furthermore, the parallel processing capabilities of neuromorphic hardware could aid biological fidelity, yielding more accurate representations of distributed bioelectric interactions across cells or tissues. Additionally, these systems offer significant energy efficiency, supporting extensive simulations without incurring high energy costs.

However, several challenges must be addressed to implement this strategy effectively. Integrating neuromorphic hardware with existing DRL frameworks involves technical complexity and may present methodological obstacles. Ensuring the accuracy of neuromorphic simulations, so that they reliably reflect biological bioelectric dynamics, remains a novel problem. Finally, access to the necessary neuromorphic hardware and the expertise to operate it is an added hurdle.

\subsection{Introduction to Do-Calculus} \label{sec:docalc}
Judea Pearl’s \textit{do-calculus} is a mathematical framework designed to infer causal relationships and predict outcomes based on interventions in complex systems. Central to Pearl’s causal inference is the concept of causal diagrams, or \textit{causal graphs}, which represent variables as nodes and causal relationships as directed edges between them. These causal diagrams enable formal analysis of causality by allowing us to reason about the effects of interventions \cite{pearl_book, Pearl1995}. The \textit{do-calculus} itself consists of rules that help transform expressions involving interventions into expressions based solely on observable quantities, thus making causal inferences feasible even from observational data \cite{Pearl2010}.

\subsubsection{Back-Door Criterion and Causal Inference}

One foundational concept in do-calculus is the \textit{back-door criterion}, a method to identify a set of variables (often referred to as \textit{back-door variables}) that must be conditioned upon to block all confounding paths between two variables. By blocking these paths, we can isolate the direct causal effect of an intervention on an outcome. Formally, for a causal effect $P(Y \mid \text{do}(X))$ to be identifiable, there must exist a set of variables $Z$ such that:
\begin{enumerate}
    \item No node in $Z$ is a descendant of $X$.
    \item $Z$ blocks every path from $X$ to $Y$ that contains an arrow pointing into $X$.
\end{enumerate}
Once the set $Z$ is identified, the causal effect of $X$ on $Y$ can be computed using observational data as:
\[
P(Y \mid \text{do}(X)) = \sum_{z} P(Y \mid X, Z = z) P(Z = z).
\]
This criterion is critical in adjusting for confounding variables and obtaining unbiased estimates of causal effects \cite{Pearl1995, Pearl2010}.

\subsubsection{Rules of Do-Calculus}

Do-calculus comprises three core rules that allow for the manipulation of probabilities involving interventions:
\begin{itemize}
    \item \textbf{Rule 1 (Insertion/Deletion of Observations)}: This rule allows us to insert or delete observations if they are conditionally independent from the outcome, given the intervention and other variables.
    \[
    P(Y \mid \text{do}(X), Z) = P(Y \mid \text{do}(X)) \quad \text{if } (Y \perp Z \mid X)
    \]
    \item \textbf{Rule 2 (Action/Observation Exchange)}: This rule enables the replacement of interventions with observations when the intervention is conditionally independent of other variables given a set of observed variables.
    \[
    P(Y \mid \text{do}(X), Z) = P(Y \mid X, Z) \quad \text{if } (Y \perp \text{do}(X) \mid Z)
    \]
    \item \textbf{Rule 3 (Insertion/Deletion of Actions)}: This rule allows us to insert or remove interventions when they do not affect the outcome due to existing dependencies.
    \[
    P(Y \mid \text{do}(X), \text{do}(Z)) = P(Y \mid \text{do}(X)) \quad \text{if } (Y \perp Z \mid X)
    \]
\end{itemize}

These rules provide a systematic approach for transforming expressions involving causal interventions, allowing researchers to determine whether a causal effect can be estimated from observational data. The rules also facilitate the derivation of expressions that capture the direct effect of an action, independent of confounding variables \cite{pearlcauses}.

\subsubsection{Application in Bioelectric and Morphogenetic Control}

In our context, integrating do-calculus within a Deep Reinforcement Learning (DRL) framework allows for the analysis of causal relationships in bioelectric signalling and cellular behaviour. By using causal diagrams, we can model the bioelectric environment and the potential effects of interventions by the DRL agent. The agent can apply interventions on bioelectric variables (e.g., ion channel states) and observe the resulting morphogenetic changes, using do-calculus to refine its understanding of the causal structure. Over time, this approach will help approximate the probability distributions of actions and outcomes, thus aiding in the development of predictive models for tissue growth, repair, and organogenesis.

\newpage

\newgeometry{left=1in, right=1in, top=1in, bottom=1in} 
\sloppy 
\bibliographystyle{plain} 
\bibliography{main}

\begin{thebibliography}{10}

\bibitem{adams2013endogenous}
David~S. Adams and Michael Levin.
\newblock Endogenous voltage gradients as mediators of cell-cell communication: strategies for investigating bioelectrical signals during pattern formation.
\newblock {\em Cell and Tissue Research}, 352(1):95--122, 4 2013.

\bibitem{Beane2013}
W.~S. Beane, J.~Morokuma, J.~M. Lemire, and M.~Levin.
\newblock Bioelectric signaling regulates head and organ size during planarian regeneration.
\newblock {\em Development (Cambridge, England)}, 140(2):313--322, 2013.

\bibitem{Beane2011-yb}
Wendy~S. Beane, Junji Morokuma, Dany~S. Adams, and Michael Levin.
\newblock A chemical genetics approach reveals h,k-atpase-mediated membrane voltage is required for planarian head regeneration.
\newblock {\em Chemistry \& Biology}, 18(1):77--89, 1 2011.

\bibitem{Benjamin2014}
Ben~Varkey Benjamin, Peiran Gao, Emmett McQuinn, Swadesh Choudhary, Anand~R. Chandrasekaran, Jean-Marie Bussat, Rodrigo Alvarez-Icaza, John~V. Arthur, Paul~A. Merolla, and Kwabena Boahen.
\newblock Neurogrid: A mixed-analog-digital multichip system for large-scale neural simulations.
\newblock {\em Proceedings of the IEEE}, 102(5):699--716, 2014.

\bibitem{Blackiston2011}
D.~Blackiston, D.~S. Adams, J.~M. Lemire, M.~Lobikin, and M.~Levin.
\newblock Transmembrane potential of glycl-expressing instructor cells induces a neoplastic-like conversion of melanocytes via a serotonergic pathway.
\newblock {\em Disease Models \& Mechanisms}, 4(1):67--85, 2011.

\bibitem{Chernet2013}
Brook~T. Chernet and Michael Levin.
\newblock Transmembrane voltage potential is an essential cellular parameter for the detection and control of tumor development in a xenopus model.
\newblock {\em Disease Models \& Mechanisms}, 6(3):595--607, 5 2013.

\bibitem{Ciaunica2023}
A.~Ciaunica, M.~Levin, F.~E. Rosas, and K.~Friston.
\newblock Nested selves: Self-organization and shared markov blankets in prenatal development in humans.
\newblock {\em Topics in Cognitive Science}, 10:10.1111/tops.12717, 2023.
\newblock Advance online publication.

\bibitem{Davies2020}
Jamie~A. Davies and Foteini Glykofrydis.
\newblock Engineering pattern formation and morphogenesis.
\newblock {\em Biochemical Society Transactions}, 48(3):1177--1185, 2020.

\bibitem{deCarvalho2024deep}
Gonçalo~H. de~Carvalho and T.~Vos.
\newblock Deep reinforcement learning in high-dimensional and stochastic environments: A short introductory literature survey \& critical methodological review.
\newblock {\em Preprint on OSF}, 2024, 8 2024.

\bibitem{scalableElectro}
Kairu Dong, Wen-Che Liu, Yuyan Su, Yidan Lyu, Hao Huang, Nenggan Zheng, John~A. Rogers, and Kewang Nan.
\newblock Scalable electrophysiology of millimeter-scale animals with electrode devices.
\newblock {\em BME Frontiers}, 4:0034, 2023.

\bibitem{pearlcauses}
Joseph~Y. Halpern and Judea Pearl.
\newblock Causes and explanations: a structural-model approach: part i: causes.
\newblock In {\em Proceedings of the Seventeenth Conference on Uncertainty in Artificial Intelligence}, UAI'01, page 194–202, San Francisco, CA, USA, 2001. Morgan Kaufmann Publishers Inc.

\bibitem{Hansali2024}
S.~Hansali, L.~Pio-Lopez, J.~V. LaPalme, and M.~Levin.
\newblock The role of bioelectrical patterns in regulative morphogenesis: An evolutionary simulation and validation in planarian regeneration.
\newblock {\em Preprints}, 2024.

\bibitem{Hartl}
Benedikt Hartl, Michael Levin, and Andreas Zöttl.
\newblock Neuroevolution of decentralized decision-making in n-bead swimmers leads to scalable and robust collective locomotion.
\newblock 2024.

\bibitem{Hartl2024}
Benjamin Hartl, Sebastian Risi, and Michael Levin.
\newblock Evolutionary implications of self-assembling cybernetic materials with collective problem-solving intelligence at multiple scales.
\newblock {\em Entropy (Basel, Switzerland)}, 26(7):532, 2024.

\bibitem{Indiveri2011}
Giacomo Indiveri, Bernabe Linares-Barranco, Tara~Julia Hamilton, Andr{\'e} van Schaik, Ralph Etienne-Cummings, Tobi Delbruck, Shih-Chii Liu, Piotr Dudek, Philipp H{\"a}fliger, Sylvie Renaud, Johannes Schemmel, Gert Cauwenberghs, John Arthur, Kai Hynna, Fopefolu Folowosele, Sylvain Saighi, Teresa Serrano-Gotarredona, Jayawan Wijekoon, Yingxue Wang, and Kwabena Boahen.
\newblock Neuromorphic silicon neuron circuits.
\newblock {\em Front Neurosci}, 5:73, 5 2011.

\bibitem{Kofman2024}
K.~Kofman and M.~Levin.
\newblock Bioelectric pharmacology of cancer: A systematic review of ion channel drugs affecting the cancer phenotype.
\newblock {\em Progress in Biophysics and Molecular Biology}, 191:25--39, 2024.

\bibitem{Lesch2010}
C.~Lesch, J.~Jo, Y.~Wu, G.~S. Fish, and M.~J. Galko.
\newblock A targeted uas-rnai screen in drosophila larvae identifies wound closure genes regulating distinct cellular processes.
\newblock {\em Genetics}, 186(3):943--957, 2010.

\bibitem{Levin2009}
M.~Levin.
\newblock Bioelectric mechanisms in regeneration: Unique aspects and future perspectives.
\newblock {\em Seminars in Cell \& Developmental Biology}, 20(5):543--556, 2009.

\bibitem{Levin2024}
M.~Levin.
\newblock Self-improvising memory: A perspective on memories as agential, dynamically reinterpreting cognitive glue.
\newblock {\em Entropy}, 26(6):481, 2024.

\bibitem{Levin2007-eb}
Michael Levin.
\newblock Gap junctional communication in morphogenesis.
\newblock {\em Progress in Biophysics and Molecular Biology}, 94(1-2):186--206, 3 2007.

\bibitem{Levin2014wh}
Michael Levin.
\newblock Molecular bioelectricity: how endogenous voltage potentials control cell behavior and instruct pattern regulation in vivo.
\newblock {\em Molecular Biology of the Cell}, 25(24):3835--3850, 12 2014.

\bibitem{Michael2019}
Michael Levin.
\newblock The computational boundary of a “self”: Developmental bioelectricity drives multicellularity and scale-free cognition.
\newblock {\em Frontiers in Psychology}, 10:2688, 12 2019.

\bibitem{Li2023}
Houpu Li, Hsin-ya Yang, Narges Asefifeyzabadi, Prabhat Baniya, Andrea~Medina Lopez, Anthony Gallegos, Kan Zhu, Hao-Chieh Hsieh, Tiffany Nguyen, Cristian Hernandez, Ksenia Zlobina, Cynthia Recendez, Maryam Tebyani, H{\'e}ctor Carri{\'o}n, John Selberg, Le~Luo, Moyasar~A. Alhamo, Athena~M. Soulika, Michael Levin, Narges Norouzi, Marcella Gomez, Min Zhao, Mircea Teodorescu, Roslyn~Rivkah Isseroff, and Marco Rolandi.
\newblock Programmable delivery of fluoxetine via wearable bioelectronics for wound healing in vivo.
\newblock {\em bioRxiv}, 2023.

\bibitem{Lobikin2012}
M.~Lobikin, B.~Chernet, D.~Lobo, and M.~Levin.
\newblock Resting potential, oncogene-induced tumorigenesis, and metastasis: the bioelectric basis of cancer in vivo.
\newblock {\em Physical Biology}, 9(6):065002, 2012.

\bibitem{Lyon2021}
P.~Lyon, F.~Keijzer, D.~Arendt, and M.~Levin.
\newblock Reframing cognition: getting down to biological basics.
\newblock {\em Philosophical Transactions of the Royal Society of London. Series B, Biological Sciences}, 376(1820):20190750, 2021.

\bibitem{Maass1997}
Wolfgang Maass.
\newblock Networks of spiking neurons: The third generation of neural network models.
\newblock {\em Neural Networks}, 10(9):1659--1671, 1997.

\bibitem{McCaig2005}
C.~D. McCaig, A.~M. Rajnicek, B.~Song, and M.~Zhao.
\newblock Controlling cell behavior electrically: current views and future potential.
\newblock {\em Physiological Reviews}, 85(3):943--978, 2005.

\bibitem{McLaughlin2018}
K.~A. McLaughlin and M.~Levin.
\newblock Bioelectric signaling in regeneration: Mechanisms of ionic controls of growth and form.
\newblock {\em Developmental Biology}, 433(2):177--189, 2018.

\bibitem{openai2019dota}
OpenAI, :, Christopher Berner, Greg Brockman, Brooke Chan, Vicki Cheung, Przemysław Dębiak, Christy Dennison, David Farhi, Quirin Fischer, Shariq Hashme, Chris Hesse, Rafal Józefowicz, Scott Gray, Catherine Olsson, Jakub Pachocki, Michael Petrov, Henrique~P. d.~O.~Pinto, Jonathan Raiman, Tim Salimans, Jeremy Schlatter, Jonas Schneider, Szymon Sidor, Ilya Sutskever, Jie Tang, Filip Wolski, and Susan Zhang.
\newblock Dota 2 with large scale deep reinforcement learning.
\newblock {\em Unknown}, page Unknown, 2019.

\bibitem{Pai2012}
V.~P. Pai, S.~Aw, T.~Shomrat, J.~M. Lemire, and M.~Levin.
\newblock Transmembrane voltage potential controls embryonic eye patterning in xenopus laevis.
\newblock {\em Development (Cambridge, England)}, 139(2):313--323, 2012.

\bibitem{Pearl1995}
Judea Pearl.
\newblock Causal diagrams for empirical research.
\newblock {\em Biometrika}, 82(4):669--688, 1995.

\bibitem{pearl_book}
Judea Pearl.
\newblock {\em Causality: Models, Reasoning and Inference}.
\newblock Cambridge University Press, USA, 2nd edition, 2009.

\bibitem{Pearl2010}
Judea Pearl.
\newblock An introduction to causal inference.
\newblock {\em The International Journal of Biostatistics}, 6(2):7, 2010.

\bibitem{Pezzulo2016}
G.~Pezzulo and M.~Levin.
\newblock Top-down models in biology: explanation and control of complex living systems above the molecular level.
\newblock {\em Journal of the Royal Society, Interface}, 13(124):20160555, 2016.

\bibitem{Pietak2016}
Alexis Pietak and Michael Levin.
\newblock Exploring instructive physiological signaling with the bioelectric tissue simulation engine (betse).
\newblock {\em Frontiers in Bioengineering and Biotechnology}, 4:55, 2016.
\newblock (Supplement).

\bibitem{PioLopez2024}
L.~Pio-Lopez and M.~Levin.
\newblock Universal multilayer network embedding reveals a causal link between gaba neurotransmitter and cancer.
\newblock 7 2024.

\bibitem{Roy2019}
Kaushik Roy, Akhilesh Jaiswal, and Priyadarshini Panda.
\newblock Towards spike-based machine intelligence with neuromorphic computing.
\newblock {\em Nature}, 575(7784):607--617, 11 2019.

\bibitem{Schrittwieser_2020}
Julian Schrittwieser, Ioannis Antonoglou, Thomas Hubert, Karen Simonyan, Laurent Sifre, Simon Schmitt, Arthur Guez, Edward Lockhart, Demis Hassabis, Thore Graepel, Timothy Lillicrap, and David Silver.
\newblock Mastering atari, go, chess and shogi by planning with a learned model.
\newblock {\em Nature}, 588(7839):604–609, December 2020.

\bibitem{SHREESHA2024150396}
Lakshwin Shreesha and Michael Levin.
\newblock Stress sharing as cognitive glue for collective intelligences: A computational model of stress as a coordinator for morphogenesis.
\newblock {\em Biochemical and Biophysical Research Communications}, 731:150396, 2024.

\bibitem{Silver2016}
David Silver, Aja Huang, Chris~J. Maddison, Arthur Guez, Laurent Sifre, George van~den Driessche, Julian Schrittwieser, Ioannis Antonoglou, Veda Panneershelvam, Marc Lanctot, Sander Dieleman, Dominik Grewe, John Nham, Nal Kalchbrenner, Ilya Sutskever, Timothy Lillicrap, Madeleine Leach, Koray Kavukcuoglu, Thore Graepel, and Demis Hassabis.
\newblock Mastering the game of go with deep neural networks and tree search.
\newblock {\em Nature}, 529(7587):484--489, 1 2016.

\bibitem{silver2017mastering}
David Silver, Thomas Hubert, Julian Schrittwieser, Ioannis Antonoglou, Matthew Lai, Arthur Guez, Marc Lanctot, Laurent Sifre, Dharshan Kumaran, Thore Graepel, Timothy Lillicrap, Karen Simonyan, and Demis Hassabis.
\newblock Mastering chess and shogi by self-play with a general reinforcement learning algorithm.
\newblock {\em Unknown}, page Unknown, 2017.

\bibitem{Sole2024}
R.~Sole, N.~Conde, J.~Pla~Mauri, J.~Garcia~Ojalvo, N.~Montserrat, and M.~Levin.
\newblock Open problems in synthetic multicellularity.
\newblock {\em Preprints}, 2024.

\bibitem{Souidi2021}
Anissa Souidi and Krzysztof Jagla.
\newblock Drosophila heart as a model for cardiac development and diseases.
\newblock {\em Cells}, 10(11):3078, 11 2021.

\bibitem{Sundelacruz2009}
S.~Sundelacruz, M.~Levin, and D.~L. Kaplan.
\newblock Role of membrane potential in the regulation of cell proliferation and differentiation.
\newblock {\em Stem Cell Reviews and Reports}, 5(3):231--246, 2009.

\bibitem{Tissot2024}
Tazzio Tissot, Mike Levin, Chris Buckley, and Richard Watson.
\newblock An ability to respond begins with inner alignment: How phase synchronisation effects transitions to higher levels of agency.
\newblock {\em bioRxiv}, 2024.

\bibitem{vinyals2019grandmaster}
Oriol Vinyals, Igor Babuschkin, Wojciech~M. Czarnecki, Micha{\"e}l Mathieu, Andrew Dudzik, Junyoung Chung, David~H. Choi, Richard Powell, Timo Ewalds, Petko Georgiev, Junhyuk Oh, Dan Horgan, Manuel Kroiss, Ivo Danihelka, Aja Huang, L.~Sifre, Trevor Cai, John~P. Agapiou, Max Jaderberg, Alexander~Sasha Vezhnevets, R{\'e}mi Leblond, Tobias Pohlen, Valentin Dalibard, David Budden, Yury Sulsky, James Molloy, Tom~Le Paine, Caglar Gulcehre, Ziyun Wang, Tobias Pfaff, Yuhuai Wu, Roman Ring, Dani Yogatama, Dario W{\"u}nsch, Katrina McKinney, Oliver Smith, Tom Schaul, Timothy~P. Lillicrap, Koray Kavukcuoglu, Demis Hassabis, Chris Apps, and David Silver.
\newblock Grandmaster level in starcraft ii using multi-agent reinforcement learning.
\newblock {\em Nature}, 575:350 -- 354, 2019.

\bibitem{Wasserthal2007}
Lutz~T Wasserthal.
\newblock Drosophila flies combine periodic heartbeat reversal with a circulation in the anterior body mediated by a newly discovered anterior pair of ostial valves and 'venous' channels.
\newblock {\em J Exp Biol}, 210(Pt 21):3707--3719, 11 2007.

\bibitem{Whited2019}
Jessica~L. Whited and Michael Levin.
\newblock Bioelectrical controls of morphogenesis: from ancient mechanisms of cell coordination to biomedical opportunities.
\newblock {\em Current Opinion in Genetics \& Development}, 57:61--69, 2019.

\bibitem{zhang2023classicalsortingalgorithmsmodel}
Taining Zhang, Adam Goldstein, and Michael Levin.
\newblock Classical sorting algorithms as a model of morphogenesis: self-sorting arrays reveal unexpected competencies in a minimal model of basal intelligence.
\newblock {\em Adaptive Behavior}, 2023.

\end{thebibliography}

\end{document}